\documentclass{article}

     \PassOptionsToPackage{numbers, compress}{natbib}

\usepackage[final]{neurips_2025}

\usepackage[utf8]{inputenc} 
\usepackage[T1]{fontenc}    
\usepackage{hyperref}       
\usepackage{url}            
\usepackage{booktabs}       
\usepackage{amsfonts}       
\usepackage{nicefrac}       
\usepackage{microtype}      
\usepackage{xcolor}         
\usepackage{amsmath}
\usepackage{amssymb}
\usepackage{graphicx}
\usepackage{subcaption}
\usepackage{float} 
\usepackage{colortbl}
\usepackage{subcaption}
\usepackage{tabularx}
\usepackage{amsmath,amssymb,amsthm,bbm}
\usepackage{multirow}
\usepackage{adjustbox}
\usepackage{algpseudocode}
\usepackage{amsmath}
\usepackage[noEnd=false]{algpseudocodex}
\usepackage[toc,page]{appendix} 
\usepackage{tocloft} 
\usepackage{titlesec}
\usepackage{booktabs}
\usepackage{threeparttable}
\usepackage{enumitem}
\setlist[enumerate]{left=10pt}
\usepackage[font=small]{caption}
\usepackage{wrapfig}
\usepackage{listings}
\usepackage{booktabs}
\usepackage{makecell}
\usepackage{algorithm}
\usepackage{algpseudocode}
\makeatletter
\newcommand\midfootnotesize{%
  \@setfontsize\midfootnotesize{8}{9}%
}
\makeatother

\usepackage{mdframed}

\newtheorem{proposition}{Proposition}
\newtheorem{lemma}{Lemma}

\definecolor{lightgreen}{rgb}{0.88,1,0.88} 

\newcounter{todocounter}
\title{Multi-Objective-Guided Discrete Flow Matching\\for Controllable Biological Sequence Design}

\author{
  Tong Chen$^{1,2}$, Yinuo Zhang$^{1, 3}$, Sophia Tang$^{1,4}$,  
  \textbf{Pranam Chatterjee}$^{1, 5, 6, \dag}$ \\\\
  $^{1}$Department of Biomedical Engineering, Duke University \\
  $^{2}$Department of Computer Science, Fudan University \\
  $^{3}$Center of Computational Biology, Duke-NUS Medical School \\
  $^{4}$Management and Technology Program, University of Pennsylvania \\
  $^{5}$Department of Computer Science, Duke University \\
  $^{6}$Department of Biostatistics and Bioinformatics, Duke University
  \\\\
  $^{\dag}$Corresponding author: 
  \href{mailto:pranam.chatterjee@duke.edu}{pranam.chatterjee@duke.edu}
}

\begin{document}

\maketitle

\begin{abstract}
\looseness=-1
Designing biological sequences that satisfy multiple, often conflicting, functional and biophysical criteria remains a central challenge in biomolecule engineering. While discrete flow matching models have recently shown promise for efficient sampling in high‐dimensional sequence spaces, existing approaches address only single objectives or require continuous embeddings that can distort discrete distributions. We present \textbf{Multi‐Objective‐Guided Discrete Flow Matching (MOG-DFM)}, a general framework to steer any pretrained discrete flow matching generator toward Pareto‐efficient trade‐offs across multiple scalar objectives. At each sampling step, MOG-DFM computes a hybrid rank‐directional score for candidate transitions and applies an adaptive hypercone filter to enforce consistent multi‐objective progression. We also trained two unconditional discrete flow matching models, \textbf{PepDFM} for diverse peptide generation and \textbf{EnhancerDFM} for functional enhancer DNA generation, as base generation models for MOG-DFM. We demonstrate MOG‑DFM’s effectiveness in generating peptide binders optimized across five properties (hemolysis, non‑fouling, solubility, half‑life, and binding affinity), and in designing DNA sequences with specific enhancer classes and DNA shapes. In total, MOG-DFM proves to be a powerful tool for multi-property-guided biomolecule sequence design. 
\end{abstract}


\section{Introduction}

Designing biological sequences that simultaneously satisfy multiple functional and biophysical criteria is a foundational challenge in modern bioengineering \citep{naseri2020application}. For example, when engineering therapeutic proteins, one must balance high target‐binding affinity with low immunogenicity and favorable pharmacokinetics \citep{tominaga2024designing}; CRISPR guide RNAs require both high on‐target activity and minimal off‐target effects \citep{mohr2016crispr, schmidt2025genome}; and synthetic promoters must achieve strong gene expression while maintaining tissue‐specific activation \citep{artemyev2024synthetic}. 

Most existing biomolecule-design methods focus on optimizing a single objective in isolation \citep{zhou2019all, nehdi2020novel}. For example, efforts have been made to reduce protein toxicity \citep{kreiser2020therapeutic, sharma2022toxinpred2} and neural networks are used to improve protein thermostability \citep{komp2025neural}. While these single‐objective approaches yield high performance on their target metrics, they often produce sequences with undesirable trade‐offs—high‐affinity peptides may be insoluble or toxic, and stabilized proteins may lose functional specificity \citep{bigi2023toxicity, rinauro2024misfolded}. Consequently, a framework for multi‐objective guided generation that can balance conflicting requirements is critical to meet the demands of practical biomolecular engineering.

Classical multi‐objective optimization (MOO) techniques, such as evolutionary algorithms and Bayesian optimization, have been successfully applied to black‐box tuning of molecular libraries \citep{zitzler1998multiobjective, deb2011multi, ueno2016combo, frisby2021bayesian}. More recently, controllable generative models have been developed to integrate MOO directly into the sampling process \citep{li2018multi, sousa2021combining, yao2024proud}. ParetoFlow \citep{yuan2024paretoflow}, for instance, leverages continuous-space flow matching to produce Pareto-optimal samples, but operates only in continuous domains. Applying such techniques to discrete sequences typically requires embedding into a continuous manifold, which can distort distributions and complicate property-based guidance \citep{beliakov2007challenges, michael2024continuous}. Recently, we introduced PepTune \citep{tang2025peptune}, a multi-objective framework based on the masked discrete diffusion language model (MDLM) architecture \cite{sahoo2024mdlm}. PepTune uses a Monte Carlo Tree Search (MCTS) strategy to guide the unmasking process toward Pareto-optimal peptide SMILES, enabling optimization across multiple therapeutic properties \citep{tang2025peptune}. However, MDLMs lack a coherent notion of token-level velocity, making them less amenable to structured, stepwise control.

Discrete flow matching has recently emerged as a powerful paradigm for directly modeling and sampling from complex discrete spaces \citep{gat2024discrete, dunn2024exploring}. Two primary variants exist: (i) continuous-time simplex methods, which diffuse discrete data through a continuous embedding over the probability simplex \citep{stark2024dirichlet, davis2024fisher, tang2025gumbel}, and (ii) jump-process models that learn time-dependent transition rates for token-level stochastic updates \citep{gat2024discrete}. The latter is particularly well suited for controllable generation, as it naturally supports reweighting of token transitions based on scalar reward functions.

Recent work has applied these models to single-objective tasks: \citet{nisonoff2025guidance} introduced rate-based classifier guidance for pretrained samplers, while \citet{tang2025gumbel} proposed Gumbel-Softmax Flow Matching with straight-through guidance for controllable discrete generation. Yet, to our knowledge, no prior work has extended discrete flow matching to support Pareto-guided generation across multiple objectives.

As such, our key contributions are as follows:
\begin{enumerate}[noitemsep,topsep=0pt,parsep=0pt,partopsep=0pt]
\item \textbf{MOG-DFM: Multi-Objective-Guided Discrete Flow Matching}, a general framework that steers pretrained discrete flow matching models toward Pareto-efficient solutions via multi-objective guidance and adaptive hypercone filtering.
\item \textbf{Rank-Directional Scoring and Hypercone Filtering} combine rank-normalized local improvement and directional alignment with a user-specified trade-off vector to reweight token-level transition velocities, followed by a dynamic angular filtering mechanism that enforces directional consistency along the Pareto front.
\item \textbf{Unconditional Base Models for Biomolecule Generation}; we train two high-quality discrete flow matching models—\textbf{PepDFM} for diverse peptide generation and \textbf{EnhancerDFM} for functional enhancer DNA generation—demonstrating low loss and biological plausibility.
\item \textbf{Multi-Property Sequence Design}; we apply MOG-DFM to two challenging biological generation tasks: (i) therapeutic peptide binder generation with five competing objectives (affinity, solubility, hemolysis, half-life, non-fouling), and (ii) enhancer DNA sequence generation guided by enhancer class and DNA shape.
\item \textbf{Superior Multi-Objective Optimization}; MOG-DFM significantly outperforms classical evolutionary algorithms and flow-based baselines on both peptide and DNA tasks, producing sequences with favorable trade-offs and improved downstream docking, folding, and property scores.
\end{enumerate}

\section{Discrete Flow Matching}
\label{overview}
In the discrete setting, we consider data $x=(x_{1},\dots,x_{d})$ taking values in a finite state space $S=\mathcal{T}^{d}$, where $\mathcal{T}=[K] = \{1,2,\dots,K\}$ is called the vocabulary. We model a continuous-time Markov chain (CTMC) $\{X_{t}\}_{t\in[0,1]}$ whose time‐dependent transition rates $u_{t}(y,x)$ transport the probability mass from an initial distribution $p_{0}$ to a target distribution $p_{1}$ \citep{gat2024discrete}. The marginal probability at time $t$ is denoted $p_{t}(x)$, and its evolution is governed by the Kolmogorov forward equation
\begin{equation}
\frac{\mathrm{d}}{\mathrm{d}t}p_{t}(y)\;=\;\sum_{x\in S}u_{t}(y,x)\,p_{t}(x)\,.
\end{equation}
The learnable velocity field $u_{t}(y,x)$ is defined as the sum of factorized velocities: 
\begin{equation}
u_t(y,x) = \displaystyle\sum_{i} \delta(y^{\bar{i}}, x^{\bar{i}})u_t^i(y^i,x),
\end{equation}
where $\bar{i} = (1,\dots,i-1,i+1,\dots,d)$ denotes all indices excluding $i$. The rate conditions for factorized velocities $u_t^i(y^i,x)$ are required per dimension $i\in[d]$: 
\begin{equation}
u_{t}(y,x)\ge 0 \; \text{for all }y^i \neq x^i, \text{ and} \;
\sum_{y^i \in \mathcal{T}}u_t^i(y^i,x)=0 \; \text{for all } x\in S,
\end{equation}
so that for small $h>0$ , the one‐step kernel
\begin{equation}
p_{t+h\mid t}(y\mid x)
=\delta(y,x)+h\,u_{t}(y,x)+o(h)
\end{equation}
remains a proper probability mass function. 

The goal of training a discrete flow matching model is to learn the velocity field $u_t^\theta$. Representing the marginal velocity $u_t^\theta$ in terms of factorized velocities $u_t^{\theta,i}$ enables the following conditional flow matching loss 
\begin{equation}
\mathcal{L}_{\mathrm{CDFM}}(\theta) = \mathbb{E}_{t, Z, X_t \sim p_{t|Z}} \sum_i D_{X_t}^i \left( u_t^i(\cdot, X_t \mid Z), u_t^{\theta, i}(\cdot, X_t) \right),
\end{equation}
where $t \sim \mathcal{U}[0,1]$, and $u_t^i(\cdot, x \mid z), u_t^{\theta, i}(\cdot, x) \in \mathbb{R}^{\mathcal{T}}$ satisfy the rate conditions. This means that $u_t^i(\cdot, x \mid z), u_t^{\theta, i}(\cdot, x) \in \Omega_{x^i}$ where, for $\alpha \in \mathcal{T}$, we define
\begin{equation}
\Omega_{\alpha} = \left\{ v \in \mathbb{R}^{\mathcal{T}} \,\middle|\, v(\beta) \geq 0 \;\forall\, \beta \in \mathcal{T} \setminus \{\alpha\}, \;\text{and}\; v(\alpha) = -\sum_{\beta \neq \alpha} v(\beta) \right\} \subset \mathbb{R}^{\mathcal{T}}.
\end{equation}

This is a convex set, and $D_x^i(u, v)$ is a Bregman divergence defined by a convex function $\Phi_x^i : \Omega_{x^i} \to \mathbb{R}$. 

In practice, we can further parameterize the velocity field using a mixture path. Specifically, a mixture path is defined with scheduler $\kappa_{t}\in[0,1]$ so that each coordinate $X_{t}^{i}$ equals $x_{0}^{i}$ or $x_{1}^{i}$ with probabilities $1-\kappa_{t}$ and $\kappa_{t}$, respectively. The mixture marginal velocity is then obtained by averaging the conditional rates over the posterior of $(x_{0},x_{1})$ given $X_{t}=x$, yielding
\begin{equation}
u^{i}_{t}(y^{i},x)
=\sum_{x_{1}^{i}}
\frac{\dot{\kappa}_{t}}{1-\kappa_{t}}
\bigl[\delta(y^{i},x_{1}^{i})-\delta(y^{i},x^{i})\bigr]\,
p^{i}_{1\mid t}(x_{1}^{i}\mid x),
\end{equation}
where $\dot{\kappa}_{t}$ denotes the time derivative of ${\kappa}_{t}$. Therefore, the aim of discrete flow matching model training, which is to learn the velocity field $u_t^i(y^i,x)$, is now equivalent to learning the marginal posterior $p^{i}_{1\mid t}(x_{1}^{i}\mid x)$. In this case, we can set the Bregman divergence to the generalized KL comparing general vectors $u,v\in\mathbb{R}_{\ge0}^{m}$,
\begin{equation}
D(u,v)
=\sum_{j}\Bigl[u_{j}\log\frac{u_{j}}{v_{j}}-u_{j}+v_{j}\Bigr].
\end{equation}
For this choice of $D$, we get 
{\small
\begin{equation}
    D\left( u_t^i(\cdot, x^i \mid x_0, x_1), u_t^{\theta, i}(\cdot, x) \right)
= \frac{\dot \kappa_t}{1 - \kappa_t} \left[ (\delta(x_1^i, x^i) - 1) \log p_{1|t}^{\theta, i}(x_1^i \mid x) + \delta(x_1^i, x^i) - p_{1|t}^{\theta, i}(x^i \mid x) \right]
\end{equation}
}

which implements the loss (8) when conditioning on $Z=(X_0, X_1)$. The generalized KL loss also provides an evidence lower bound (ELBO) on the likelihood of the target distribution
\begin{equation}
- \log p_1^{\theta}(x_1) \le \mathbb{E}_{t, X_0, X_t \sim p_{t \mid 0,1}} \sum_i D\left( u_t^i(\cdot, X_1^i \mid X_0, x_1), u_t^{\theta, i}(\cdot, X_t) \right),
\end{equation}
where $p_1^\theta$ is the marginal generated by the model at time $t=1$. Therefore, in addition to training, the generalized KL loss can also be used for evaluation. 

\begin{figure}
    \centering
    \includegraphics[width=\textwidth]{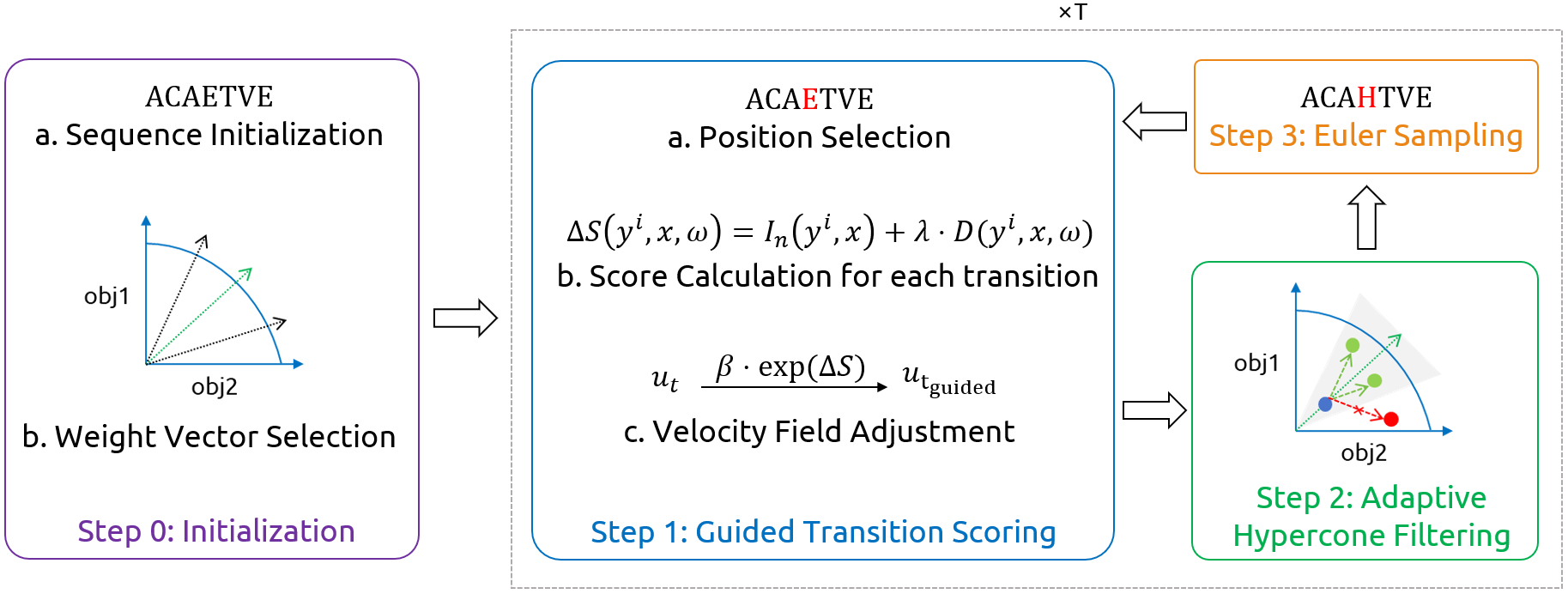}
    \caption{Visualization for MOG-DFM algorithm.}
    \label{fig:mog-dfm}
    \vskip -0.1in
\end{figure}

\section{Multi-Objective Guided Discrete Flow Matching} 
MOG-DFM (\textbf{M}ulti-\textbf{O}bjective \textbf{G}uided \textbf{D}iscrete \textbf{F}low \textbf{M}atching) operates under the same setting as discrete flow matching described in the previous section. Suppose we have a pre‑trained discrete flow matching model that defines a CTMC with a factorized velocity field $u_t^i(y^i,x)$ , which transports probability mass from an initial distribution $p_{0}$ to the unknown target distribution via mixture path parametrization. In addition, we assume access to $N$ pre-trained scalar score functions $s_{n}:\mathcal{S}\to\mathbb{R},\text{where} \ n=1,\dots,N$, that assign objective scores to any sequence.  Our aim is to generate novel sequences $x_{1}\in\mathcal{S}$ whose objective vectors $\bigl(s_{1}(x_{1}),\,s_{2}(x_{1}),\,\dots,s_{N}(x_{1})\bigr)$ lie near the Pareto front (not guaranteed to be Pareto optimal)
\[
\mathrm{PF} \;=\; \bigl\{x\in\mathcal{S}\,\bigm|\,\nexists\,x'\in\mathcal{S}:\;s_{n}(x')\ge s_{n}(x)\;\forall n,\;\exists\,m:s_{m}(x')>s_{m}(x)\bigr\}.
\]
To achieve this, we will guide the CTMC sampling dynamics of the discrete flow matching model using multi‑objective transition scores, steering the generative process toward Pareto‑efficient regions of the state space (Figure \ref{fig:mog-dfm}, Pseudocode \ref{algorithm}, Proof in Section \ref{proof}). 

\subsection{Step 0: Initialization and Weight Vector Generation}
MOG-DFM begins by initializing the generative process at time $t=0$ by sampling an initial sequence $x_{0}$ uniformly from the discrete state space $\mathcal{S}=[K]^{d}$.  To steer the generation towards diverse Pareto‐efficient solutions, we introduce a set of $M$ weight vectors $\{\omega^{k}\}_{k=1}^{M}$, where $\omega\in \mathbb{R}^N$, that uniformly cover the $N$‐dimensional Pareto front.  Intuitively, each $\omega$ encodes a particular trade‐off among the $N$ objectives, so sampling different $\omega$ promotes exploration of distinct regions of the Pareto front.  We construct these vectors via the Das–Dennis simplex lattice with $H$ subdivisions, yielding components
\begin{equation}
    \omega_{i}=\frac{k_{i}}{H},\quad k_{i}\in\mathbb{Z}_{\ge0},\quad\sum_{i=1}^{N}k_{i}=H,
\end{equation}
and then draw one $\omega$ randomly before the following steps. This defines one direction we want to optimize toward the Pareto front in the current run. The following three steps will then be performed in each iteration. We set the number of total iterations to $T$.

\subsection{Step 1: Guided Transition Scoring}
We first randomly select one position $i$ in the sequence that will be updated during the current iteration. At each intermediate state $x_{t}$ and selected position $i$, each possible candidate transition $y^{i}\neq x^{i}$ is scored by combining local improvement measures with global directional alignment. The normalized rank score captures how much each individual objective improves relative to other possible token replacements, thereby encouraging exploration of promising local moves. Formally, for each objective $n$ we compute
\begin{equation}
    I_{n}(y^{i},x)
=\frac{\mathrm{rank}\bigl(s_{n}(x_{\mathrm{new}})-s_{n}(x)\bigr)}{|T|},
\end{equation}
where  $x_{\mathrm{new}}$ denotes the sequence obtained by replacing the $i$th token of $x$ with $y^{i}$. The $\text{rank}(\cdot)$ function maps the raw score change into a uniform scale in $[0,1]$.  In contrast, the directional term
\begin{equation}
    D(y^{i},x,\omega)
=\Delta\mathbf{s}(y^{i},x)\cdot\omega
\end{equation}
measures the alignment of the multi‐objective improvement vector $\Delta\mathbf{s}$ with the chosen weight vector $\omega$, ensuring that transitions not only improve individual objectives but collectively move toward the desired trade‐off direction. To balance rank‐based exploration against direction‐guided exploitation, we z‐score normalize both components and combine them as follows
\begin{equation}
    \Delta S(y^{i},x,\omega)
=\mathrm{Norm}\Bigl[\tfrac{1}{N}\sum_{n=1}^{N}i_{n}I_{n}(y^{i},x)\Bigr]
+\lambda\,\mathrm{Norm}\bigl[D(y^{i},x,\omega)\bigr],
\end{equation}
where $\lambda>0$ is a tunable hyperparameter.  An importance vector $\mathbf{I} = [i_1,\dots,i_N]$ is used to normalize the improvement values for each objective. Finally, we re-weight the original factorized velocity field from the pre-trained discrete flow matching model:
\begin{equation}
    u_{\mathrm{guided},t}^{\,i}(y^{i},x\mid\omega)
= \begin{cases}
\beta \, u_{t}^{\,i}(y^{i},x)\,\exp\bigl(\Delta S(y^{i},x,\omega)\bigr)&
\quad y^{i}\neq x^{i} \\
-\sum_{y^{i}\neq x^{i}}u_{\mathrm{guided},t}^{\,i}(y^{i},x\mid\omega)& \quad y^{i} =x^{i}
\end{cases}
\end{equation}

where $\beta$  is the strength hyperparameter. Therefore, the guided velocities satisfy the non-negativity and zero‐sum rate conditions by construction, preserving valid CTMC dynamics while favoring high‐utility transitions. 

\subsection{Step 2: Adaptive Hypercone Filtering}

To ensure each candidate token replacement drives the sequence towards the chosen trade‑off direction, we restrict candidate transitions to lie within a cone around the weight vector \(\omega\).  This “hypercone” mechanism allows the sampler to navigate non‑convex or discontinuous regions of the Pareto front by enforcing local directional consistency. Specifically, for a given position \(i\) and candidate token \(y^{i}\), we compute the angle
\begin{equation}
    \alpha^{i}
=\arccos\biggl(\frac{\Delta\mathbf{s}(y^{i},x)\,\cdot\,\omega}{\lVert\Delta\mathbf{s}(y^{i},x)\rVert\;\lVert\omega\rVert}\biggr),
\end{equation}
where \(\Delta\mathbf{s}(y^{i},x)\) is the multi‑objective improvement vector from replacing \(x^{i}\) with \(y^{i}\).  We accept only those \(y^{i}\) for which $\alpha^{i}\le\Phi$, where \(\Phi\) denotes the current hypercone angle. Denoting \(Y^{i}\subseteq T\setminus\{x^{i}\}\) as the set of accepted tokens, we select the best transition as 
\begin{equation}
    y_{\mathrm{best}}^{i}
=\arg\max_{\,y^{i}\in Y^{i}}
\Delta S(y^{i},x,\omega) \quad \text{if} \; Y^i \neq \emptyset.
\end{equation}
There are two degenerate cases that can lead to empty $Y^i$: (1) if every \(\alpha^{i}\ge\pi\), indicating that all possible transitions decrease performance, we will perform a self‑transition and retain the current state; (2) if there exist some \(\alpha^{i}<\pi\) but none lie within the cone (i.e.,\ \(\Phi\) is temporarily too small), we still advance by choosing the best‑aligned candidate 
\begin{equation}
    y_{\mathrm{best}}^{i}
=\arg\max_{\{y':\alpha^{i}<\pi\}}
\Delta S(y^{i},x,\omega),
\end{equation}
allowing progress while the hypercone angle self‑adjusts.

As a pre-defined hypercone angle may be too big or too small during the dynamic optimization process, we need to adaptively tune the angle that best balances exploration and exploitation. Specifically, we compute the rejection rate
\begin{equation}
    r_{t}
=\frac{\#\{y^{i}:\alpha^{i}>\Phi\}}
{\text{total \# of candidate transitions}}
\end{equation}
and its exponential moving average (EMA)
\begin{equation}
    \bar r_{t}
=\alpha_{r}\,\bar r_{t-h}
+\bigl(1-\alpha_{r}\bigr)\,r_{t},
\end{equation}
where \(\alpha_{r}\in[0,1)\) is a smoothing coefficient and \(\bar r_{0}=\tau\) is the target rejection rate. We then update the hypercone angle via
\begin{equation}
    \Phi_{t+h}
=\mathrm{clip}\Bigl(
\Phi_{t}\,\exp\bigl(\eta\,(\bar r_{t}-\tau)\bigr),\,
\Phi_{\min},\,\Phi_{\max}
\Bigr),
\end{equation}
with learning rate \(\eta>0\) and bounds \(\Phi_{\min},\Phi_{\max}\) to prevent the hypercone from collapsing or over‑expanding.  Intuitively, if too many candidates are being rejected (\(\bar r_{t}>\tau\)), the hypercone widens to admit more directions; if too few are rejected (\(\bar r_{t}<\tau\)), it narrows to focus on the most aligned transitions.  

\subsection{Step 3: Euler Sampling}

Once the guided transition rates \(u_{\mathrm{guided},t}^{\,i}(y^{i},x\mid\omega)\) have been computed and the best candidate transition has been selected after hypercone filtering (if not self-transitioning), we evolve the CTMC via Euler sampling. Specifically, we denote the total outgoing rate from \(x\) at time \(t\) on coordinate \(i\) by
\begin{equation}
    R_{t}^{\,i}(x)
=-\,u_{\mathrm{guided},t}^{\,i}(x^{i},x\mid\omega)
=\sum_{y^{i}\neq x^{i}}u_{\mathrm{guided},t}^{\,i}(y^{i},x\mid\omega).
\end{equation}
The one‐step transition kernel for coordinate \(i\) is given by the exact Euler–Maruyama analogue for CTMCs:
\begin{equation}
\mathbb{P}\bigl(X^{i}_{t+h}=y^{i}\mid X_{t}=x\bigr)
=
\begin{cases}
\exp\bigl(h\,u_{\mathrm{guided},t}^{\,i}(x^{i},x\mid\omega)\bigr)
=\exp\bigl(-h\,R_{t}^{\,i}(x)\bigr), 
&y^{i}=x^{i},\\[6pt]
\dfrac{u_{\mathrm{guided},t}^{\,i}(y^{i},x\mid\omega)}{R_{t}^{\,i}(x)}\,
\bigl(1-\exp(-h\,R_{t}^{\,i}(x))\bigr),
&y^{i}\neq x^{i}.
\end{cases}
\end{equation}
Here, $h = 1/T$ is the step size in the time interval, $X_t$ and $X_{t+h}$ denotes the current state and the next state respectively. In practice, one draws a uniform random number \(r\in[0,1]\): if \(r \le 1- \exp(-h\,R_{t}^{\,i}(x))\), $x^i$ will transition to the best selected candidate; otherwise, we retain $x^i$. 

After performing step 1 to step 3 for $T$ iterations, we end with the final sample $x_1$ whose score vectors have been steered close to the Pareto front, with all objectives optimized. 

\section{Experiments}
To the best of our knowledge, there are no public datasets that serve to benchmark multi-objective optimization algorithms for biological sequences. Therefore, we develop two benchmarks to evaluate MOG-DFM: multi-objective guided peptide binder sequence generation and multi-objective guided enhancer DNA sequence generation. We first show two discrete flow matching models developed for peptide generation and enhancer DNA generation, and we then demonstrate MOG-DFM's efficacy on a wide variety of tasks and examples. 

\subsection{PepDFM and EnhancerDFM Generate Diverse and Biologically Plausible Sequences}
To enable the efficient generation of peptide binders, we developed an unconditional peptide generator, \textbf{PepDFM}, based on the Discrete Flow Matching (DFM) framework. The model backbone of PepDFM is a U-Net-style convolutional architecture. We trained PepDFM on a custom dataset that includes all peptides from the PepNN and BioLip2 datasets, as well as sequences from the PPIRef dataset with lengths ranging from 6 to 49 amino acids, finally converging to a training loss of 3.3134 and a validation loss of 3.1051 \cite{abdin2022pepnn, zhang2024biolip2, bushuiev2023learning}. As described in Section \ref{overview}, the low generalized KL loss during evaluation demonstrates the strong performance of PepDFM. We further investigate the diversity and biological plausibility of peptides generated by PepDFM. Specifically, PepDFM generates peptides with substantially high Hamming distances from the test set, indicating a great degree of diversity and novelty in the generated sequences (Figure \ref{fig:shannon}). Additionally, the Shannon entropy of the generated peptides closely matches that of the test set, highlighting the model's capability to produce biologically plausible peptides with diverse sequence lengths (Figure \ref{fig:shannon}).

\renewcommand{\arraystretch}{1.3}
\begin{table}[t]
\centering
\caption{\textbf{Evaluation of unconditional EnhancerDNA generation.} Each method generates 10k sequences, and we compare their empirical distributions with the data distributions using the Fr\'echet Biological distance (FBD) metric. NFE refers to number of function evaluations. \# Training Epochs refers to the number of training epochs needed to get the model checkpoint for this evaluation. The Random Sequence baseline shows the FBD for the same number and length of sequences with uniform randomly chosen nucleotides. Dirichlet FM refers to the Dirichlet Flow Matching model \citep{stark2024dirichlet} }
\vspace{0.5em}
\label{tab:enhancerdfm}
\begin{tabular}{l|c|c|c}
\hline
\textbf{} & \textbf{FBD} & \textbf{NFE} & \textbf{\# Training Epochs} \\
\hline
Random Sequence & 622.8 & - & - \\
Dirichlet FM    & 5.3   & 100 & 1400 \\
EnhancerDFM     & 5.9   & 100 & 20\\
\hline
\end{tabular}
\end{table}

EnhancerDFM adopts the same model backbone and melanoma enhancer dataset used in the enhancer DNA design task from Stark, et al. \cite{stark2024dirichlet}. We employed the Fr\'echet Biological distance (FBD) metric from \cite{stark2024dirichlet} to evaluate the performance of EnhancerDFM (Table \ref{tab:enhancerdfm}). Specifically, using the same number of function evaluations (NFE), EnhancerDFM achieved a comparable FBD of 5.9 compared with Dirichlet FM of 5.3, significantly lower than the FBD of random sequences, demonstrating EnhancerDFM's ability to design biologically plausible enhancer DNA sequences. Significantly, the best EnhancerDFM model is achieved within 20 training epochs, while the best EnhancerDFM is obtained only in around 1400 training epochs, highlighting discrete flow matching models' superior capability of capturing the underlying data distribution. 

\subsection{MOG-DFM effectively balances each objective trade-off}
To validate that the MOG-DFM framework can balance the trade-offs between each objective, we performed two sets of experiments for peptide binder generation with three property guidance, and in ablation experiment settings, we removed one or more objectives. In the binder design task for target 7LUL (affinity, solubility, hemolysis guidance; Table~\ref{tab:ablation_7LUL}), omitting any single guidance causes a collapse in that property, while the remaining guided metrics may modestly improve. Likewise, in the binder design task for target CLK1 (affinity, non‑fouling, half‑life guidance; Table~\ref{tab:ablation_CLK1}), disabling non‑fouling guidance allows half‑life to exceed 80 hours but drives non‑fouling near zero, and disabling half‑life guidance preserves non‑fouling yet reduces half‑life below 2 hours. In contrast, enabling all guidance signals produces the most balanced profiles across all objectives. These results confirm that MOG‑DFM precisely targets chosen objectives while preserving the flexibility to navigate conflicting requirements and push samples toward the Pareto front, thereby demonstrating the correctness and precision of our multi‑objective sampling framework.

\renewcommand{\arraystretch}{1.3}
\begin{table}[t]
\midfootnotesize
\centering
\caption{MOG-DFM generates peptide binders for 10 diverse protein targets, optimizing five therapeutic properties: hemolysis, non-fouling, solubility, half-life (in hours), and binding affinity. Each value represents the average of 100 MOG-DFM-designed binders.}
\vspace{0.5em}
\begin{tabular}{c|c|c|c|c|c|c}
\hline
\textbf{Target} & \makecell{\textbf{Binder}\\\textbf{Length}} & \textbf{Hemolysis ($\downarrow$)} & \textbf{Non-Fouling ($\uparrow$)}& \textbf{Solubility ($\uparrow$)}& \textbf{Half-Life ($\uparrow$)}& \textbf{Affinity ($\uparrow$)} \\
\hline
AMHR2       & 8  & 0.0755 & 0.8352 & 0.8219 & 31.624 & 7.3789 \\
AMHR2       & 12 & 0.0570 & 0.8419 & 0.8279 & 28.761 & 7.4274 \\
AMHR2       & 16 & 0.0618 & 0.7782 & 0.7428 & 31.227 & 7.6099 \\
EWS::FLI1    & 8  & 0.0809 & 0.8508 & 0.8296 & 47.169 & 6.2251 \\
EWS::FLI1    & 12 & 0.0616 & 0.8302 & 0.8130 & 34.225 & 6.3631 \\
EWS::FLI1    & 16 & 0.0709 & 0.7787 & 0.7400 & 34.192 & 6.5912 \\
MYC         & 8  & 0.0809 & 0.8135 & 0.8005 & 39.836 & 6.8488 \\
OX1R        & 10 & 0.0741 & 0.8115 & 0.7969 & 33.533 & 7.4162 \\
DUSP12      & 9  & 0.0735 & 0.8360 & 0.8216 & 33.754 & 6.4946 \\
 1B8Q& 8& 0.0744& 0.8334& 0.827& 33.243&5.932\\
 1E6I        & 6  & 0.0887 & 0.7884 & 0.7793 & 41.164 &4.9621 \\
3IDJ& 7& 0.0924& 0.8246& 0.7992& 30.388& 7.6304\\
5AZ8        & 11 & 0.0698 & 0.8462 & 0.8420 & 28.726 & 6.6051 \\
7JVS        & 11 & 0.0628 & 0.8390 & 0.8206 & 32.834 & 6.9569 \\
\hline

\end{tabular}

\label{tab:5_classifiers}
\end{table}

\subsection{MOG-DFM generates peptide binders under five property guidance}
We next benchmark MOG-DFM on a peptide binder generation task guided by five different properties that are critical for therapeutic discovery: hemolysis, non-fouling, solubility, half-life, and binding affinity. To evaluate MOG‑DFM in a controlled setting, we designed 100 peptide binders per target for ten diverse proteins—structured targets with known binders (1B8Q, 1E6I, 3IDJ, 5AZ8, 7JVS), structured targets without known binders (AMHR2, OX1R, DUSP12), and intrinsically disordered targets (EWS::FLI1, MYC) (Table~\ref{tab:5_classifiers}). Across all targets and across multiple binder lengths, the generated peptides achieve low hemolysis rates (0.06–0.09), high non‑fouling ($>$0.78) and solubility ($>$0.74), extended half‑life (28–47 h), and strong affinity scores (6.4–7.6), demonstrating both balanced optimization and robustness to sequence length. 

\begin{figure}
    \centering
    \includegraphics[width=\textwidth]{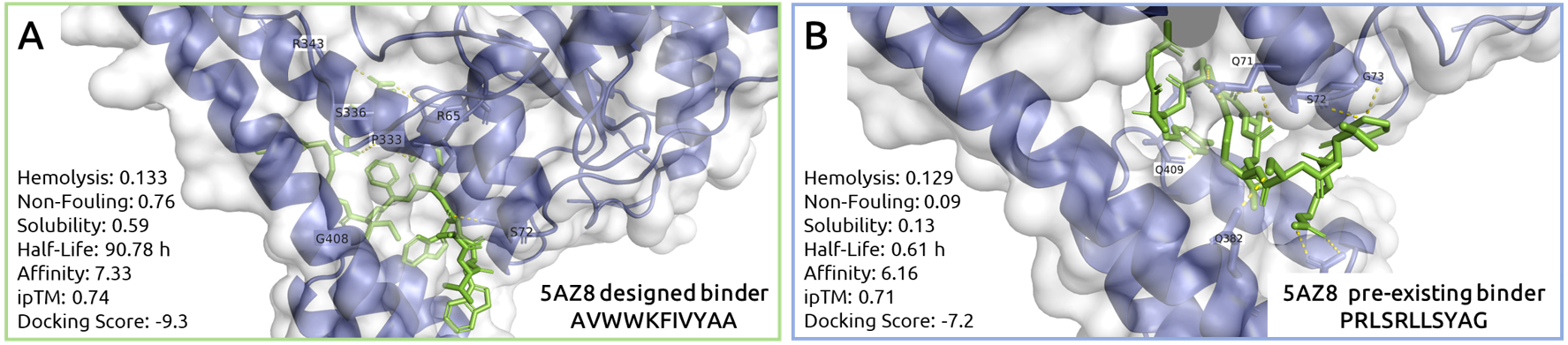}
    \vspace{0.3em}
      \includegraphics[width=\textwidth]{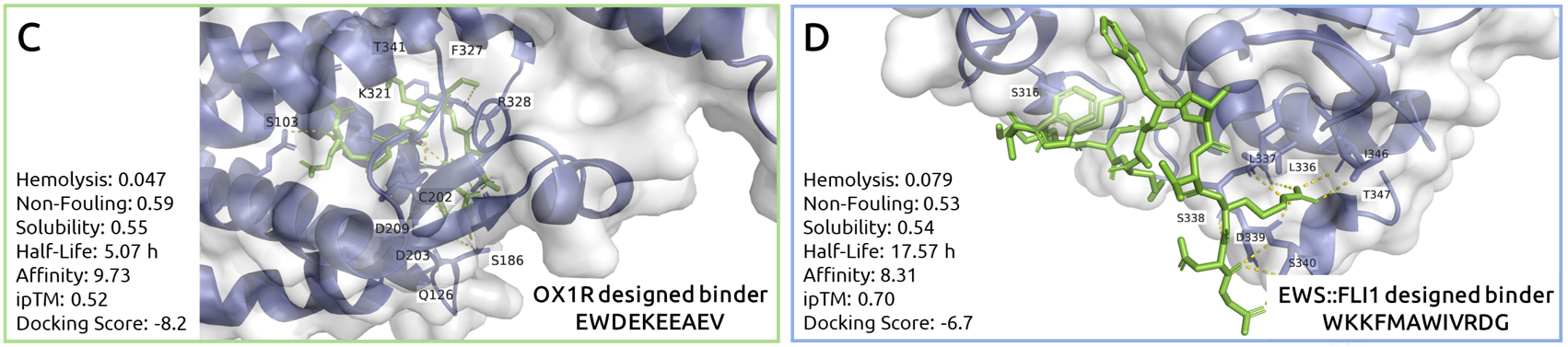}
          \vspace{0.3em}
      \includegraphics[width=\textwidth]{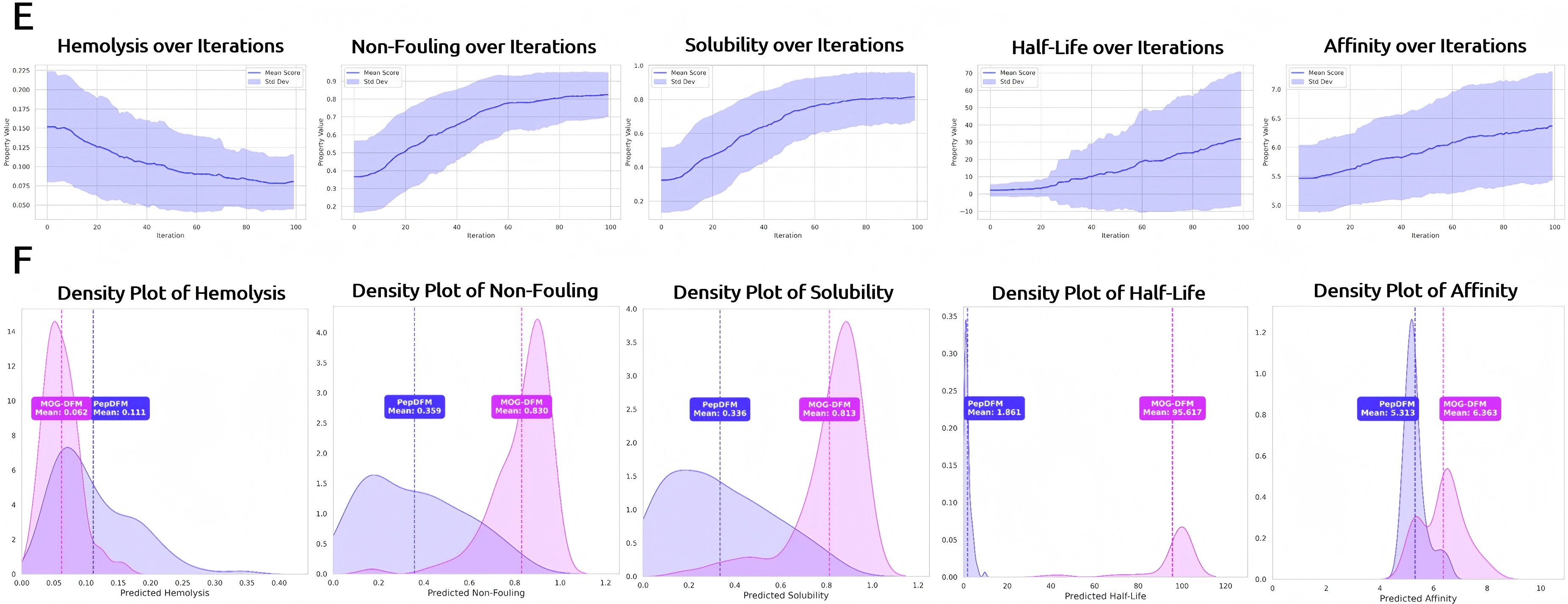}
    \caption{\textbf{(A), (B)} Complex structures of PDB 5AZ8 with a MOG-DFM-designed binder and its pre-existing binder. \textbf{(C), (D)} Complex structures of two target proteins without pre-existing binders (OX1R, EWS::FLI1) with MOG-DFM-designed binders. Five property scores are shown for each binder, along with the ipTM score from AlphaFold3 and docking score from AutoDock VINA. Interacting residues on the target are visualized. \textbf{(E)} Plots showing the mean scores for each property across the number of iterations during MOG-DFM's design of binders of length 12-aa for EWS::FLI1. \textbf{(F)} Density plots illustrating the distribution of predicted property scores for MOG-DFM-designed EWS::FLI1 binders  of length 12 aa, compared to the peptides generated unconditionally by PepDFM. Please zoom in for better viewing.}
    \label{fig:visualization}
    \vskip -0.1in
\end{figure}

For the target proteins with pre-existing binders, we compared the property values between their known binders with MOG-DFM-designed ones (Figure \ref{fig:visualization}A,B, \ref{fig:w_binders}). The designed binders significantly outperform the pre-existing binders across all properties without compromising the binding potential, which is further confirmed by the ipTM scores computed by AlphaFold3 \citep{abramson2024accurate} and docking scores calculated by AutoDock VINA \citep{trott2010autodock}. Although the MOG-DFM-designed binders bind to similar target positions as the pre-existing ones, they differ significantly in sequence and structure, demonstrating MOG-DFM’s capacity to explore the vast sequence space for optimal designs. For target proteins without known binders, complex structures were visualized using one of the MOG-DFM-designed binders (Figure \ref{fig:visualization}C,D, \ref{fig:w/o_binders}). The corresponding property scores, as well as ipTM and docking scores, are also displayed. Some of the designed binders demonstrated extended half-life, while others excelled in non-fouling and solubility, underscoring the comprehensive exploration of the sequence space by MOG-DFM.

At each iteration, we recorded the mean and standard deviation of the five property scores across all the 100 binders to evaluate the effectiveness of the guided generation strategy (Figure \ref{fig:visualization}E). All five properties exhibited an improving trend over iterations, with the average score of the solubility and non-fouling properties showing a significant increase from score around 0.3 to 0.8. A large deviation of the final half-life values is caused by the susceptibility of the half-life value to guidance, with MOG-DFM balancing the trade-offs between half-life and other values. The improvements of hemolysis, non-fouling, and solubility gradually converge, demonstrating MOG-DFM's efficiency in steering the generation process to the Pareto Front within only 100 iterations.

We visualized the distribution change steered by MOG-DFM by plotting the property score distribution of 100 peptides of length 12 designed for EWS::FLI1 and 100 peptides of the same length sampled unconditionally from PepDFM (Figure \ref{fig:visualization}F). MOG-DFM effectively shifted and concentrated the peptide distribution so that the peptides possess improved properties for all the objectives, demonstrating MOG-DFM's ability to steer the generation so that all properties are optimized simultaneously. 

In Section \ref{score_models}, we demonstrate the reliability of our score models. We now use external evaluation tools to further confirm that MOG-DFM-designed binders possess desired properties. The average solubility and half‑life for each target across all 100 designed peptides were predicted using ADMET-AI (Table~\ref{tab:admet_ai}) \citep{swanson2024admet}.  ADMET‑AI, trained on a different dataset from our solubility and half-life prediction models, predicts average LogS values around –2.5 $\text{log mol·L}^{-1}$, which is well above the conventional –4 threshold for good solubility, and confirms long half‑life estimates (>15 h).  These results from an orthogonal predictive model demonstrate MOG‑DFM's capability to generate candidates with multiple desirable drug properties. 

\begin{table}[t]
\midfootnotesize
\centering
\caption{MOG-DFM outperforms traditional multi-objective optimization algorithms in designing peptide binders guided by five objectives. Each value represents the average of 100 designed binders. The table also records the average runtime for each algorithm to design a single binder. The best result for each metric is highlighted in bold.}
\label{tab:comparison}
\vspace{0.5em}
\begin{tabular}{cccccccc}
\hline
\textbf{Target} & \textbf{Method}   & \textbf{Time (s)} & \textbf{Hemolysis ($\downarrow$)} & \textbf{Non‑Fouling} & \textbf{Solubility} & \textbf{Half‑Life} & \textbf{Affinity} \\
\hline
\multirow{5}{*}{1B8Q} 
  & MOPSO     & 8.54& 0.1066& 0.4763& 0.4684& 4.449& 6.0594\\
  & NSGA‑III  &  33.13& 0.0862& 0.5715& 0.5825&  7.324& 7.2178
\\
  & SMS‑EMOA  &  \textbf{8.21} & 0.1196& 0.3450& 0.3511&  3.023& 5.955
\\
  & SPEA2     &  17.48& 0.0819& 0.4973& 0.5057&  4.126& \textbf{7.324}
\\
  & \textbf{MOG-DFM} &  43.00& \textbf{0.0785} & \textbf{0.8445} & \textbf{0.8455} &  \textbf{27.227} & 5.9094 \\
\hline
\multirow{5}{*}{PPP5} 
  & MOPSO& 11.34& 0.0883& 0.4711& 0.4255& 1.769& 6.6958\\
  & NSGA‑III  & 37.30& \textbf{0.0479} & 0.7138& 0.7066&  2.901& 7.3789
\\
  & SMS‑EMOA  &  \textbf{8.43} & 0.1242& 0.4269& 0.4334&  1.031& 6.2854
\\
  & SPEA2     &  19.02& 0.0555& 0.6221& 0.6098& 2.613& \textbf{7.6253}
\\
  & \textbf{MOG‑DFM}   &  90.00& 0.0617& \textbf{0.7738} & \textbf{0.751} &  \textbf{27.775} & 6.8197
\\
\hline
\end{tabular}
\end{table}

We benchmarked MOG‑DFM against four multi‑objective optimizers—NSGA‑III \citep{deb2013evolutionary}, SMS‑EMOA \citep{beume2007sms}, SPEA2 \citep{zitzler2001spea2}, and MOPSO \citep{coello2002mopso}—on two protein targets: 1B8Q (a small protein with known peptide binders) and PPP5 (a larger protein lacking characterized binders) (Table~\ref{tab:comparison}). For each method, we generated 100 peptide binders per target of a specified length, guided by five property objectives (hemolysis, non‑fouling, solubility, half‑life, and binding affinity), and recorded both the average generation time for one sequence and the mean property scores. Although MOG‑DFM requires longer runtimes, it consistently produces the most favorable trade-offs: reducing predicted hemolysis by more than 10\%, boosting non‑fouling and solubility by approximately 30-50\%, and extending half‑life by a factor of 3 to 4 compared to the next‑best method, while maintaining competitive affinity values. These results demonstrate MOG‑DFM’s effectiveness in navigating high‑dimensional property landscapes to generate peptide binders with well-balanced, optimized profiles. We did not benchmark against ParetoFlow, another multi-objective optimization algorithm that uses flow matching, because it requires score models to take continuous inputs, which is not suitable for our task \citep{yuan2024paretoflow}. 

\subsection{Hyperparameter Sensitivity Benchmark}
There are several hyperparameters in MOG‑DFM whose settings may affect generative performance. To assess this sensitivity, we evaluated peptide binder design across a broad range of values for each parameter (Table~\ref{tab:hyperparam_sensitivity}). We find that increasing the number of sampling steps consistently improves all performance metrics, as finer discretization more closely approximates the continuous‐time dynamics. In contrast, setting the initial hypercone angle \(\Phi_{\mathrm{init}}\) too small or too large both degrade results: an overly narrow cone restricts exploration, while an overly wide cone dilutes directional guidance. The importance weights also play a critical role in balancing multiple objectives. Because each property can vary over a different numerical range, we initialize each weight inversely proportional to the maximum observed improvement of that property, thereby normalizing all guidance signals to roughly unit scale. This allows for similar improvements for each objective, otherwise the improvements for some objectives may stagnate. By comparison, the remaining hyperparameters (i.e., $\beta$, \(\lambda\), \(\alpha_{r}\), \(\eta\), \(\tau\), and the bounds \(\Phi_{\min},\Phi_{\max}\)) exhibit only modest impact on outcomes, indicating that MOG‑DFM is robust to moderate variations in these settings.  

\subsection{Adaptive Hypercone Filtering Enhances Multi-Objective Optimization}
To quantify the contribution of our adaptive hypercone mechanism, we performed an ablation study on three increasingly disordered, and thus undruggable, protein targets (3IDJ, 4E‑BP2, and EWS::FLI1), generating 100 peptide binders for each target (Table~\ref{tab:algorithm_ablation}).  Removing hypercone filtering entirely (“w/o filtering”) causes a dramatic collapse in half‑life—from roughly 30–35 hours down to 4–13 hours—while leaving non‑fouling and solubility largely unchanged, indicating that filtering out poorly aligned moves is essential for optimizing objectives that require gradual, coordinated changes.  Introducing static hypercone gating without angle adaptation (“w/o adaptation”) recovers much of the half‑life gains (to 23–37 h), but at the expense of reduced non‑fouling and solubility scores and only marginal improvements in affinity.  In contrast, the full MOG‑DFM—with both directional hypercone filtering and adaptive angle updates—simultaneously elevates half‑life and maintains strong performance across all five objectives.  This effect is especially pronounced on disordered targets (4E‑BP2 and EWS::FLI1), where dynamic cone adjustment is essential for navigating the irregular, non‑convex Pareto landscapes.  

\begin{table}[ht]
\centering
\vspace{0.5em}
\caption{\textbf{Performance evaluation of MOG-DFM in guided DNA sequence generation.} Task 1 guides the generation towards the HelT shape and enhancer class 1, while Task 2 targets the Rise shape and enhancer class 16. The table presents the predicted DNA shape values (HelT for Task 1, Rise for Task 2) and enhancer class probabilities (class 1 for Task 1, class 16 for Task 2) under various guidance conditions. The 'Shape' column shows the predicted DNA shape values obtained using Deep DNAshape, and the 'Class Prob' column displays the predicted enhancer class probabilities. Ablation studies were conducted by removing one or both guidance criteria, as shown by the rows corresponding to different combinations of shape and class guidance. For each setting, 5 enhancer DNA sequences were designed.}
\vspace{0.5 em}
\label{tab:dna}
\begin{tabular}{|cc|cc|cc|}
\hline
\multicolumn{2}{|c|}{\textbf{Guidance Settings}} & \multicolumn{2}{c|}{\textbf{Task 1}} & \multicolumn{2}{c|}{\textbf{Task 2}} \\
\hline
\textbf{Shape} & \textbf{Class} & \textbf{Class Prob} & \textbf{Shape} & \textbf{Class Prob} & \textbf{Shape} \\
\hline

\multirow{5}{*}{\checkmark} & \multirow{5}{*}{\checkmark} & 0.7504 & 36.0100 & 0.9960 & 3.3640 \\
 & & 0.6507 & 36.0100 & 0.9922 & 3.3680 \\
 & & 0.6821 & 36.0000 & 0.9864 & 3.3669 \\
 & & 0.7097 & 36.0000 & 0.9976 & 3.3680 \\
 & & 0.6425 & 36.0000 & 0.9961 & 3.3623 \\
\hline

\multirow{5}{*}{$\checkmark$} & \multirow{5}{*}{$\times$} 
& 0.9999 & 34.3274 & 1.0000 & 3.3368 \\
& & 0.9999 & 34.4715 & 1.0000 & 3.3345 \\
& & 0.9989 & 34.4257 & 0.9999 & 3.3348 \\
& & 0.9997 & 34.5226 & 0.9994 & 3.3357 \\
& & 0.9998 & 34.4210 & 1.0000 & 3.3340 \\
\hline

\multirow{5}{*}{$\times$} & \multirow{5}{*}{$\checkmark$}
& 0.0026 & 36.0017 & 2.36E-05 & 3.3690 \\
& & 0.0055 & 36.0238 & 0.0005 & 3.3647 \\
& & 0.0062 & 36.0214 & 0.0114 & 3.3705 \\
& & 0.0186 & 36.0396 & 0.0001 & 3.3717 \\
& & 0.0051 & 36.0304 & 0.0054 & 3.3669 \\
\hline

\multirow{5}{*}{$\times$} & \multirow{5}{*}{$\times$}
& 0.0362 & 34.7379 & 0.0008 & 3.3283 \\
& & 0.0364 & 34.5350 & 0.0057 & 3.3258 \\
& & 0.0309 & 34.5720 & 0.0476 & 3.3268 \\
& & 0.0138 & 34.3060 & 0.0632 & 3.3378 \\
& & 0.0213 & 34.5500 & 0.0003 & 3.3320 \\
\hline

\end{tabular}
\end{table}

\subsection{MOG-DFM generates enhancer DNA of specific class with specified DNA shapes}

To demonstrate the universal capability of MOG-DFM in performing multi-objective guided generation for biological sequences, we applied MOG-DFM to design enhancer DNA sequences guided by enhancer class and DNA shape. EnhancerDFM was used as the unconditional enhancer DNA sequence generator, while Deep DNAshape was employed to predict DNA shape \citep{li2024predicting}, and the enhancer class predictor from which it was sourced \cite{stark2024dirichlet}. Two distinct tasks with different enhancer class and DNA shape guidance were carried out, and ablation results are presented in Table \ref{tab:dna}. Given the time constraints, we designed five enhancer sequences of length 100 for each setting. 

In the first task, we conditioned the generation to target enhancer class 1 (associated with the transcription factor binding motif ATF) and a high HelT (helix twist) value, with the maximum HelT value set to 36. With both guidance criteria in place, MOG-DFM effectively steered the sequence generation towards enhancer class 1 while simultaneously ensuring that the HelT value approached its maximum (Table \ref{tab:dna}). When one or both guidance criteria were removed, the corresponding properties showed significant degradation, with the probability of achieving the desired enhancer class dropping near zero (Table \ref{tab:dna}). A similar outcome was observed in the second task, which targeted enhancer class 16 and a higher Rise shape value, with the maximum Rise value set to 3.7. Since the canonical range for the Rise shape value spans from 3.3 to 3.4, MOG-DFM ensured both a high probability for the target enhancer class and an optimal DNA shape value, outperforming other ablation settings (Table \ref{tab:dna}).


\section{Conclusion}

In this work, we have presented \textbf{Multi-Objective-Guided Discrete Flow Matching (MOG-DFM)}, a scalable framework for generating biomolecular sequences that simultaneously optimize multiple, often conflicting properties. By guiding discrete flow matching models with multi-objective optimization, MOG-DFM enables the design of peptide and DNA sequences with improved therapeutic and structural characteristics.

While MOG-DFM performs well in biological domains, future work will extend the framework to longer sequences and higher-dimensional outputs, including applications in text and image generation. From a theoretical perspective, improving Pareto convergence guarantees and incorporating uncertainty-aware or feedback-driven guidance remain key directions to explore. Ultimately, MOG-DFM offers a foundation for generating the next generation of therapeutics—molecules that are not only effective but explicitly optimized for the multifaceted properties critical to clinical success.

\section{Declarations}
\textbf{Acknowledgments.} We thank the Duke Compute Cluster, Pratt School of Engineering IT department, and Mark III Systems, for providing database and hardware support that has contributed to the research reported within this manuscript. 

\textbf{Author Contributions.} T.C. devised model architectures and theoretical formulations, and trained and benchmarked models. Y.Z. trained and benchmarked supervised models, and performed molecular docking. Y.Z. and S.T. advised on model design and theoretical framework. T.C. drafted the manuscript, and Y.Z. and S.T. assisted in figure design and data presentation. P.C. designed, supervised, and directed the study, formulated algorithm proofs, and finalized the manuscript.

\textbf{Data and Materials Availability.} The codebase will be freely accessible to the academic community at \url{https://huggingface.co/ChatterjeeLab/MOG-DFM}.

\textbf{Funding Statement.} This research was supported by grants from the Hartwell Foundation, CHDI Foundation, and EndAxD Foundation to the lab of P.C.

\textbf{Competing Interests.} P.C. is a co-founder of Gameto, Inc. and UbiquiTx, Inc. and advises companies involved in biologics development. P.C.’s interests are reviewed and managed by Duke University in accordance with their conflict-of-interest policies. T.C., S.T., and Y.Z., have no conflicts of interest to declare.






\bibliographystyle{unsrtnat}
\bibliography{citation}  

\begin{thebibliography}{53}
\providecommand{\natexlab}[1]{#1}
\providecommand{\url}[1]{\texttt{#1}}
\expandafter\ifx\csname urlstyle\endcsname\relax
  \providecommand{\doi}[1]{doi: #1}\else
  \providecommand{\doi}{doi: \begingroup \urlstyle{rm}\Url}\fi

\bibitem[Naseri and Koffas(2020)]{naseri2020application}
Gita Naseri and Mattheos~AG Koffas.
\newblock Application of combinatorial optimization strategies in synthetic biology.
\newblock \emph{Nature communications}, 11\penalty0 (1):\penalty0 2446, 2020.

\bibitem[Tominaga et~al.(2024)Tominaga, Shima, Nozaki, Ito, Someda, Shoya, Hashii, Obata, Matsumoto-Kitano, Suematsu, et~al.]{tominaga2024designing}
Masahiro Tominaga, Yoko Shima, Kenta Nozaki, Yoichiro Ito, Masataka Someda, Yuji Shoya, Noritaka Hashii, Chihiro Obata, Miho Matsumoto-Kitano, Kohei Suematsu, et~al.
\newblock Designing strong inducible synthetic promoters in yeasts.
\newblock \emph{Nature Communications}, 15\penalty0 (1):\penalty0 10653, 2024.

\bibitem[Mohr et~al.(2016)Mohr, Hu, Ewen-Campen, Housden, Viswanatha, and Perrimon]{mohr2016crispr}
Stephanie~E Mohr, Yanhui Hu, Benjamin Ewen-Campen, Benjamin~E Housden, Raghuvir Viswanatha, and Norbert Perrimon.
\newblock Crispr guide rna design for research applications.
\newblock \emph{The FEBS journal}, 283\penalty0 (17):\penalty0 3232--3238, 2016.

\bibitem[Schmidt et~al.(2025)Schmidt, Zhang, Chakarov, Bansal, Mourelatos, S{\'a}nchez-Rivera, Lowe, Ventura, Leslie, and Pritykin]{schmidt2025genome}
Henri Schmidt, Minsi Zhang, Dimitar Chakarov, Vineet Bansal, Haralambos Mourelatos, Francisco~J S{\'a}nchez-Rivera, Scott~W Lowe, Andrea Ventura, Christina~S Leslie, and Yuri Pritykin.
\newblock Genome-wide crispr guide rna design and specificity analysis with guidescan2.
\newblock \emph{Genome biology}, 26\penalty0 (1):\penalty0 1--25, 2025.

\bibitem[Artemyev et~al.(2024)Artemyev, Gubaeva, Paremskaia, Dzhioeva, Deviatkin, Feoktistova, Mityaeva, and Volchkov]{artemyev2024synthetic}
Valentin Artemyev, Anna Gubaeva, Anastasiia~Iu Paremskaia, Amina~A Dzhioeva, Andrei Deviatkin, Sofya~G Feoktistova, Olga Mityaeva, and Pavel~Yu Volchkov.
\newblock Synthetic promoters in gene therapy: Design approaches, features and applications.
\newblock \emph{Cells}, 13\penalty0 (23):\penalty0 1963, 2024.

\bibitem[Zhou et~al.(2019)Zhou, Jiang, Yang, Yu, Feng, Adil, Deng, Zou, Zhang, Lu, et~al.]{zhou2019all}
Ruimin Zhou, Zhaoyan Jiang, Chen Yang, Jianwei Yu, Jirui Feng, Muhammad~Abdullah Adil, Dan Deng, Wenjun Zou, Jianqi Zhang, Kun Lu, et~al.
\newblock All-small-molecule organic solar cells with over 14\% efficiency by optimizing hierarchical morphologies.
\newblock \emph{Nature communications}, 10\penalty0 (1):\penalty0 5393, 2019.

\bibitem[Nehdi et~al.(2020)Nehdi, Samman, Aguilar-S{\'a}nchez, Farah, Yurdusev, Boudjelal, and Perreault]{nehdi2020novel}
Atef Nehdi, Nosaibah Samman, Vanessa Aguilar-S{\'a}nchez, Azer Farah, Emre Yurdusev, Mohamed Boudjelal, and Jonathan Perreault.
\newblock Novel strategies to optimize the amplification of single-stranded dna.
\newblock \emph{Frontiers in Bioengineering and Biotechnology}, 8:\penalty0 401, 2020.

\bibitem[Kreiser et~al.(2020)Kreiser, Wright, Block, Hollows, Nguyen, LeForte, Mannini, Vendruscolo, and Limbocker]{kreiser2020therapeutic}
Ryan~P Kreiser, Aidan~K Wright, Natalie~R Block, Jared~E Hollows, Lam~T Nguyen, Kathleen LeForte, Benedetta Mannini, Michele Vendruscolo, and Ryan Limbocker.
\newblock Therapeutic strategies to reduce the toxicity of misfolded protein oligomers.
\newblock \emph{International journal of molecular sciences}, 21\penalty0 (22):\penalty0 8651, 2020.

\bibitem[Sharma et~al.(2022)Sharma, Naorem, Jain, and Raghava]{sharma2022toxinpred2}
Neelam Sharma, Leimarembi~Devi Naorem, Shipra Jain, and Gajendra~PS Raghava.
\newblock Toxinpred2: an improved method for predicting toxicity of proteins.
\newblock \emph{Briefings in bioinformatics}, 23\penalty0 (5):\penalty0 bbac174, 2022.

\bibitem[Komp et~al.(2025)Komp, Phillips, Lee, Fallin, Alanzi, Zorman, McCully, and Beck]{komp2025neural}
Evan Komp, Christian Phillips, Lauren~M Lee, Shayna~M Fallin, Humood~N Alanzi, Marlo Zorman, Michelle~E McCully, and David~AC Beck.
\newblock Neural network conditioned to produce thermophilic protein sequences can increase thermal stability.
\newblock \emph{Scientific Reports}, 15\penalty0 (1):\penalty0 14124, 2025.

\bibitem[Bigi et~al.(2023)Bigi, Lombardo, Cascella, and Cecchi]{bigi2023toxicity}
Alessandra Bigi, Eva Lombardo, Roberta Cascella, and Cristina Cecchi.
\newblock The toxicity of protein aggregates: new insights into the mechanisms, 2023.

\bibitem[Rinauro et~al.(2024)Rinauro, Chiti, Vendruscolo, and Limbocker]{rinauro2024misfolded}
Dillon~J Rinauro, Fabrizio Chiti, Michele Vendruscolo, and Ryan Limbocker.
\newblock Misfolded protein oligomers: Mechanisms of formation, cytotoxic effects, and pharmacological approaches against protein misfolding diseases.
\newblock \emph{Molecular Neurodegeneration}, 19\penalty0 (1):\penalty0 20, 2024.

\bibitem[Zitzler and Thiele(1998)]{zitzler1998multiobjective}
Eckart Zitzler and Lothar Thiele.
\newblock Multiobjective optimization using evolutionary algorithms—a comparative case study.
\newblock In \emph{International conference on parallel problem solving from nature}, pages 292--301. Springer, 1998.

\bibitem[Deb(2011)]{deb2011multi}
Kalyanmoy Deb.
\newblock Multi-objective optimisation using evolutionary algorithms: an introduction.
\newblock In \emph{Multi-objective evolutionary optimisation for product design and manufacturing}, pages 3--34. Springer, 2011.

\bibitem[Ueno et~al.(2016)Ueno, Rhone, Hou, Mizoguchi, and Tsuda]{ueno2016combo}
Tsuyoshi Ueno, Trevor~David Rhone, Zhufeng Hou, Teruyasu Mizoguchi, and Koji Tsuda.
\newblock Combo: An efficient bayesian optimization library for materials science.
\newblock \emph{Materials discovery}, 4:\penalty0 18--21, 2016.

\bibitem[Frisby and Langmead(2021)]{frisby2021bayesian}
Trevor~S Frisby and Christopher~James Langmead.
\newblock Bayesian optimization with evolutionary and structure-based regularization for directed protein evolution.
\newblock \emph{Algorithms for Molecular Biology}, 16\penalty0 (1):\penalty0 13, 2021.

\bibitem[Li et~al.(2018)Li, Zhang, and Liu]{li2018multi}
Yibo Li, Liangren Zhang, and Zhenming Liu.
\newblock Multi-objective de novo drug design with conditional graph generative model.
\newblock \emph{Journal of cheminformatics}, 10:\penalty0 1--24, 2018.

\bibitem[Sousa et~al.(2021)Sousa, Correia, Pereira, and Rocha]{sousa2021combining}
Tiago Sousa, Jo{\~a}o Correia, Vitor Pereira, and Miguel Rocha.
\newblock Combining multi-objective evolutionary algorithms with deep generative models towards focused molecular design.
\newblock In \emph{Applications of Evolutionary Computation: 24th International Conference, EvoApplications 2021, Held as Part of EvoStar 2021, Virtual Event, April 7--9, 2021, Proceedings 24}, pages 81--96. Springer, 2021.

\bibitem[Yao et~al.(2024)Yao, Pan, Li, Tsang, and Yao]{yao2024proud}
Yinghua Yao, Yuangang Pan, Jing Li, Ivor Tsang, and Xin Yao.
\newblock Proud: Pareto-guided diffusion model for multi-objective generation.
\newblock \emph{Machine Learning}, 113\penalty0 (9):\penalty0 6511--6538, 2024.

\bibitem[Yuan et~al.(2024)Yuan, Chen, Pal, and Liu]{yuan2024paretoflow}
Ye~Yuan, Can Chen, Christopher Pal, and Xue Liu.
\newblock Paretoflow: Guided flows in multi-objective optimization.
\newblock \emph{arXiv preprint arXiv:2412.03718}, 2024.

\bibitem[Beliakov and Lim(2007)]{beliakov2007challenges}
Gleb Beliakov and Kieran~F Lim.
\newblock Challenges of continuous global optimization in molecular structure prediction.
\newblock \emph{European journal of operational research}, 181\penalty0 (3):\penalty0 1198--1213, 2007.

\bibitem[Michael et~al.(2024)Michael, Bartels, Gonz{\'a}lez-Duque, Zainchkovskyy, Frellsen, Hauberg, and Boomsma]{michael2024continuous}
Richard Michael, Simon Bartels, Miguel Gonz{\'a}lez-Duque, Yevgen Zainchkovskyy, Jes Frellsen, S{\o}ren Hauberg, and Wouter Boomsma.
\newblock A continuous relaxation for discrete bayesian optimization.
\newblock \emph{arXiv preprint arXiv:2404.17452}, 2024.

\bibitem[Tang et~al.(2025{\natexlab{a}})Tang, Zhang, and Chatterjee]{tang2025peptune}
Sophia Tang, Yinuo Zhang, and Pranam Chatterjee.
\newblock Peptune: De novo generation of therapeutic peptides with multi-objective-guided discrete diffusion.
\newblock \emph{Proceedings of the 42nd International Conference on Machine Learning (ICML)}, 2025{\natexlab{a}}.

\bibitem[Sahoo et~al.(2024)Sahoo, Arriola, Schiff, Gokaslan, Marroquin, Chiu, Rush, and Kuleshov]{sahoo2024mdlm}
Subham~Sekhar Sahoo, Marianne Arriola, Yair Schiff, Aaron Gokaslan, Edgar Marroquin, Justin~T Chiu, Alexander Rush, and Volodymyr Kuleshov.
\newblock Simple and effective masked diffusion language models.
\newblock \emph{Advances in Neural Information Processing Systems}, 2024.

\bibitem[Gat et~al.(2024)Gat, Remez, Shaul, Kreuk, Chen, Synnaeve, Adi, and Lipman]{gat2024discrete}
Itai Gat, Tal Remez, Neta Shaul, Felix Kreuk, Ricky~TQ Chen, Gabriel Synnaeve, Yossi Adi, and Yaron Lipman.
\newblock Discrete flow matching.
\newblock \emph{Advances in Neural Information Processing Systems}, 37:\penalty0 133345--133385, 2024.

\bibitem[Dunn and Koes(2024)]{dunn2024exploring}
Ian Dunn and David~Ryan Koes.
\newblock Exploring discrete flow matching for 3d de novo molecule generation.
\newblock \emph{ArXiv}, pages arXiv--2411, 2024.

\bibitem[Stark et~al.(2024)Stark, Jing, Wang, Corso, Berger, Barzilay, and Jaakkola]{stark2024dirichlet}
Hannes Stark, Bowen Jing, Chenyu Wang, Gabriele Corso, Bonnie Berger, Regina Barzilay, and Tommi Jaakkola.
\newblock Dirichlet flow matching with applications to dna sequence design.
\newblock \emph{Proceedings of the 41st International Conference on Machine Learning (ICML)}, 2024.

\bibitem[Davis et~al.(2024)Davis, Kessler, Petrache, Ceylan, Bronstein, and Bose]{davis2024fisher}
Oscar Davis, Samuel Kessler, Mircea Petrache, Ismail Ceylan, Michael Bronstein, and Joey Bose.
\newblock Fisher flow matching for generative modeling over discrete data.
\newblock \emph{Advances in Neural Information Processing Systems}, 37:\penalty0 139054--139084, 2024.

\bibitem[Tang et~al.(2025{\natexlab{b}})Tang, Zhang, Tong, and Chatterjee]{tang2025gumbel}
Sophia Tang, Yinuo Zhang, Alexander Tong, and Pranam Chatterjee.
\newblock Gumbel-softmax flow matching with straight-through guidance for controllable biological sequence generation.
\newblock \emph{arXiv preprint arXiv:2503.17361}, 2025{\natexlab{b}}.

\bibitem[Nisonoff et~al.(2025)Nisonoff, Xiong, Allenspach, and Listgarten]{nisonoff2025guidance}
Hunter Nisonoff, Junhao Xiong, Stephan Allenspach, and Jennifer Listgarten.
\newblock Unlocking guidance for discrete state-space diffusion and flow models.
\newblock \emph{Proceedings of the 13th International Conference on Learning Representations (ICLR)}, 2025.

\bibitem[Abdin et~al.(2022)Abdin, Nim, Wen, and Kim]{abdin2022pepnn}
Osama Abdin, Satra Nim, Han Wen, and Philip~M Kim.
\newblock Pepnn: a deep attention model for the identification of peptide binding sites.
\newblock \emph{Communications biology}, 5\penalty0 (1):\penalty0 503, 2022.

\bibitem[Zhang et~al.(2024)Zhang, Zhang, Freddolino, and Zhang]{zhang2024biolip2}
Chengxin Zhang, Xi~Zhang, Peter~L Freddolino, and Yang Zhang.
\newblock Biolip2: an updated structure database for biologically relevant ligand--protein interactions.
\newblock \emph{Nucleic Acids Research}, 52\penalty0 (D1):\penalty0 D404--D412, 2024.

\bibitem[Bushuiev et~al.(2023)Bushuiev, Bushuiev, Kouba, Filkin, Gabrielova, Gabriel, Sedlar, Pluskal, Damborsky, Mazurenko, et~al.]{bushuiev2023learning}
Anton Bushuiev, Roman Bushuiev, Petr Kouba, Anatolii Filkin, Marketa Gabrielova, Michal Gabriel, Jiri Sedlar, Tomas Pluskal, Jiri Damborsky, Stanislav Mazurenko, et~al.
\newblock Learning to design protein-protein interactions with enhanced generalization.
\newblock \emph{arXiv preprint arXiv:2310.18515}, 2023.

\bibitem[Abramson et~al.(2024)Abramson, Adler, Dunger, Evans, Green, Pritzel, Ronneberger, Willmore, Ballard, Bambrick, et~al.]{abramson2024accurate}
Josh Abramson, Jonas Adler, Jack Dunger, Richard Evans, Tim Green, Alexander Pritzel, Olaf Ronneberger, Lindsay Willmore, Andrew~J Ballard, Joshua Bambrick, et~al.
\newblock Accurate structure prediction of biomolecular interactions with alphafold 3.
\newblock \emph{Nature}, 630\penalty0 (8016):\penalty0 493--500, 2024.

\bibitem[Trott and Olson(2010)]{trott2010autodock}
Oleg Trott and Arthur~J Olson.
\newblock Autodock vina: improving the speed and accuracy of docking with a new scoring function, efficient optimization, and multithreading.
\newblock \emph{Journal of computational chemistry}, 31\penalty0 (2):\penalty0 455--461, 2010.

\bibitem[Swanson et~al.(2024)Swanson, Walther, Leitz, Mukherjee, Wu, Shivnaraine, and Zou]{swanson2024admet}
Kyle Swanson, Parker Walther, Jeremy Leitz, Souhrid Mukherjee, Joseph~C Wu, Rabindra~V Shivnaraine, and James Zou.
\newblock Admet-ai: a machine learning admet platform for evaluation of large-scale chemical libraries.
\newblock \emph{Bioinformatics}, 40\penalty0 (7):\penalty0 btae416, 2024.

\bibitem[Deb and Jain(2013)]{deb2013evolutionary}
Kalyanmoy Deb and Himanshu Jain.
\newblock An evolutionary many-objective optimization algorithm using reference-point-based nondominated sorting approach, part i: solving problems with box constraints.
\newblock \emph{IEEE transactions on evolutionary computation}, 18\penalty0 (4):\penalty0 577--601, 2013.

\bibitem[Beume et~al.(2007)Beume, Naujoks, and Emmerich]{beume2007sms}
Nicola Beume, Boris Naujoks, and Michael Emmerich.
\newblock Sms-emoa: Multiobjective selection based on dominated hypervolume.
\newblock \emph{European journal of operational research}, 181\penalty0 (3):\penalty0 1653--1669, 2007.

\bibitem[Zitzler et~al.(2001)Zitzler, Laumanns, and Thiele]{zitzler2001spea2}
Eckart Zitzler, Marco Laumanns, and Lothar Thiele.
\newblock Spea2: Improving the strength pareto evolutionary algorithm.
\newblock \emph{TIK report}, 103, 2001.

\bibitem[Coello and Lechuga(2002)]{coello2002mopso}
CA~Coello Coello and Maximino~Salazar Lechuga.
\newblock Mopso: A proposal for multiple objective particle swarm optimization.
\newblock In \emph{Proceedings of the 2002 Congress on Evolutionary Computation. CEC'02 (Cat. No. 02TH8600)}, volume~2, pages 1051--1056. IEEE, 2002.

\bibitem[Li et~al.(2024)Li, Chiu, and Rohs]{li2024predicting}
Jinsen Li, Tsu-Pei Chiu, and Remo Rohs.
\newblock Predicting dna structure using a deep learning method.
\newblock \emph{Nature communications}, 15\penalty0 (1):\penalty0 1243, 2024.

\bibitem[Ronneberger et~al.(2015)Ronneberger, Fischer, and Brox]{ronneberger2015u}
Olaf Ronneberger, Philipp Fischer, and Thomas Brox.
\newblock U-net: Convolutional networks for biomedical image segmentation.
\newblock In \emph{Medical image computing and computer-assisted intervention--MICCAI 2015: 18th international conference, Munich, Germany, October 5-9, 2015, proceedings, part III 18}, pages 234--241. Springer, 2015.

\bibitem[Lin et~al.(2023)Lin, Akin, Rao, Hie, Zhu, Lu, Smetanin, Verkuil, Kabeli, Shmueli, et~al.]{lin2023evolutionary}
Zeming Lin, Halil Akin, Roshan Rao, Brian Hie, Zhongkai Zhu, Wenting Lu, Nikita Smetanin, Robert Verkuil, Ori Kabeli, Yaniv Shmueli, et~al.
\newblock Evolutionary-scale prediction of atomic-level protein structure with a language model.
\newblock \emph{Science}, 379\penalty0 (6637):\penalty0 1123--1130, 2023.

\bibitem[Atak et~al.(2021)Atak, Taskiran, Demeulemeester, Flerin, Mauduit, Minnoye, Hulselmans, Christiaens, Ghanem, Wouters, et~al.]{atak2021interpretation}
Zeynep~Kalender Atak, Ibrahim~Ihsan Taskiran, Jonas Demeulemeester, Christopher Flerin, David Mauduit, Liesbeth Minnoye, Gert Hulselmans, Valerie Christiaens, Ghanem-Elias Ghanem, Jasper Wouters, et~al.
\newblock Interpretation of allele-specific chromatin accessibility using cell state--aware deep learning.
\newblock \emph{Genome research}, 31\penalty0 (6):\penalty0 1082--1096, 2021.

\bibitem[Buenrostro et~al.(2013)Buenrostro, Giresi, Zaba, Chang, and Greenleaf]{buenrostro2013transposition}
Jason~D Buenrostro, Paul~G Giresi, Lisa~C Zaba, Howard~Y Chang, and William~J Greenleaf.
\newblock Transposition of native chromatin for fast and sensitive epigenomic profiling of open chromatin, dna-binding proteins and nucleosome position.
\newblock \emph{Nature methods}, 10\penalty0 (12):\penalty0 1213--1218, 2013.

\bibitem[Zhang et~al.(2023)Zhang, Wu, Xiu, Li, Chen, Wang, Wang, Gao, and Zhou]{zhang2023pepland}
Ruochi Zhang, Haoran Wu, Yuting Xiu, Kewei Li, Ningning Chen, Yu~Wang, Yan Wang, Xin Gao, and Fengfeng Zhou.
\newblock Pepland: a large-scale pre-trained peptide representation model for a comprehensive landscape of both canonical and non-canonical amino acids.
\newblock \emph{arXiv preprint arXiv:2311.04419}, 2023.

\bibitem[Guntuboina et~al.(2023)Guntuboina, Das, Mollaei, Kim, and Barati~Farimani]{guntuboina2023peptidebert}
Chakradhar Guntuboina, Adrita Das, Parisa Mollaei, Seongwon Kim, and Amir Barati~Farimani.
\newblock Peptidebert: A language model based on transformers for peptide property prediction.
\newblock \emph{The Journal of Physical Chemistry Letters}, 14\penalty0 (46):\penalty0 10427--10434, 2023.

\bibitem[Pedregosa et~al.(2011)Pedregosa, Varoquaux, Gramfort, Michel, Thirion, Grisel, Blondel, Prettenhofer, Weiss, Dubourg, Vanderplas, Passos, Cournapeau, Brucher, Perrot, and Duchesnay]{scikit-learn}
F.~Pedregosa, G.~Varoquaux, A.~Gramfort, V.~Michel, B.~Thirion, O.~Grisel, M.~Blondel, P.~Prettenhofer, R.~Weiss, V.~Dubourg, J.~Vanderplas, A.~Passos, D.~Cournapeau, M.~Brucher, M.~Perrot, and E.~Duchesnay.
\newblock Scikit-learn: Machine learning in {P}ython.
\newblock \emph{Journal of Machine Learning Research}, 12:\penalty0 2825--2830, 2011.

\bibitem[Akiba et~al.(2019)Akiba, Sano, Yanase, Ohta, and Koyama]{akiba2019optuna}
Takuya Akiba, Shotaro Sano, Toshihiko Yanase, Takeru Ohta, and Masanori Koyama.
\newblock Optuna: A next-generation hyperparameter optimization framework.
\newblock In \emph{International Conference on Knowledge Discovery and Data Mining}, pages 2623--2631, 2019.

\bibitem[Mathur et~al.(2016)Mathur, Prakash, Anand, Kaur, Agrawal, Mehta, Kumar, Singh, and Raghava]{mathur2016peplife}
Deepika Mathur, Satya Prakash, Priya Anand, Harpreet Kaur, Piyush Agrawal, Ayesha Mehta, Rajesh Kumar, Sandeep Singh, and Gajendra~PS Raghava.
\newblock Peplife: a repository of the half-life of peptides.
\newblock \emph{Scientific reports}, 6\penalty0 (1):\penalty0 36617, 2016.

\bibitem[D’Aloisio et~al.(2021)D’Aloisio, Dognini, Hutcheon, and Coxon]{d2021peptherdia}
Vera D’Aloisio, Paolo Dognini, Gillian~A Hutcheon, and Christopher~R Coxon.
\newblock Peptherdia: database and structural composition analysis of approved peptide therapeutics and diagnostics.
\newblock \emph{Drug Discovery Today}, 26\penalty0 (6):\penalty0 1409--1419, 2021.

\bibitem[Jain et~al.(2024)Jain, Gupta, Patiyal, and Raghava]{jain2024thpdb2}
Shipra Jain, Srijanee Gupta, Sumeet Patiyal, and Gajendra~PS Raghava.
\newblock Thpdb2: compilation of fda approved therapeutic peptides and proteins.
\newblock \emph{Drug Discovery Today}, page 104047, 2024.

\bibitem[Tsuboyama et~al.(2023)Tsuboyama, Dauparas, Chen, Laine, Mohseni~Behbahani, Weinstein, Mangan, Ovchinnikov, and Rocklin]{tsuboyama2023mega}
Kotaro Tsuboyama, Justas Dauparas, Jonathan Chen, Elodie Laine, Yasser Mohseni~Behbahani, Jonathan~J Weinstein, Niall~M Mangan, Sergey Ovchinnikov, and Gabriel~J Rocklin.
\newblock Mega-scale experimental analysis of protein folding stability in biology and design.
\newblock \emph{Nature}, 620\penalty0 (7973):\penalty0 434--444, 2023.

\end{thebibliography}
\appendix
\newpage

\section{Base Model Details}
\subsection{PepDFM}
\textbf{Model Architecture.} The base model is a time-dependent architecture based on U-Net \citep{ronneberger2015u}. It uses two separate embedding layers for sequence and time, followed by five convolutional blocks with varying dilation rates to capture temporal dependencies, while incorporating time-conditioning through dense layers.  The final output layer generates logits for each token. We used a polynomial convex schedule with a polynomial exponent of 2.0 for the mixture discrete probability path in the discrete flow matching. 

\textbf{Dataset Curation.} The dataset for PepDFM training was curated from the PepNN, BioLip2, and PPIRef dataset \citep{abdin2022pepnn, zhang2024biolip2, bushuiev2023learning}. All peptides from PepNN and BioLip2 were included, along with sequences from PPIRef ranging from 6 to 49 amino acids in length. The dataset was divided into training, validation, and test sets at an 80/10/10 ratio.

\textbf{Training Strategy.} The training is conducted on a 2xH100 NVIDIA NVL GPU system with 94 GB of VRAM for 200 epochs with batch size 512. The model checkpoint with the lowest evaluation loss was saved. The Adam optimizer was employed with a learning rate 1e-4. A learning rate scheduler with 20 warm-up epochs and cosine decay was used, with initial and minimum learning rates both 1e-5. The embedding dimension and hidden dimension were set to be 512 and 256 respectively for the base model.

\textbf{Dynamic Batching.} To enhance computational efficiency and manage variable-length token sequences, we implemented dynamic batching. Drawing inspiration from ESM-2’s approach \citep{lin2023evolutionary}, input peptide sequences were sorted by length to optimize GPU memory utilization, with a maximum token size of 100 per GPU.

\subsection{EnhancerDFM}
\textbf{Model Architecture.} The base model for EnhancerDFM applies the same architecture as the PepDFM. We also used a polynomial convex schedule with a polynomial exponent of 2.0 for the mixture discrete probability path in the discrete flow matching. 

\textbf{Dataset Curation.} The dataset for EnhancerDFM training is curated by \citep{stark2024dirichlet}. The dataset contains 89k enhancer sequences from human melanoma cells \citep{atak2021interpretation}. Each sequence is of length 500 paired with cell class labels determined from ATAC-seq data \citep{buenrostro2013transposition}. There are 47 such classes of cells in total, with details displayed in Table \ref{tab:enhancer_data} \citep{atak2021interpretation}. We applied the same dataset split strategy as \citep{stark2024dirichlet}.

\textbf{Training Strategy.} The training is conducted on a 2xH100 NVIDIA NVL GPU system with 94 GB of VRAM for 1500 epochs with batch size 256. The model checkpoint with the lowest evaluation loss was saved. The Adam optimizer was employed with a learning rate 1e-3. A learning rate scheduler with 150 warm-up epochs and cosine decay was used, with initial and minimum learning rates both 1e-4. Both the embedding dimension and hidden dimension were set to be 256 for the base model.

\section{Score Model Details}
\label{score_models}
We collected hemolysis (9,316), non-fouling (17,185), solubility (18,453), and binding affinity (1,781) data for classifier training from the PepLand and PeptideBERT datasets \citep{zhang2023pepland, guntuboina2023peptidebert}. All sequences taken are wild-type L-amino acids and are tokenized and represented by ESM-2 protein language model \cite{lin2023evolutionary}. 

\subsection{Boosted Trees for Classification}
For hemolysis, non-fouling, and solubility classification, we trained XGBoost boosted tree models for logistic regression. We split the data into 0.8/0.2 train/validation using traitified splits from scikit-learn \cite{scikit-learn} and generated mean pooled ESM-2-650M \cite{lin2023evolutionary} embeddings as input features to the model. We ran 50 trials of OPTUNA \cite{akiba2019optuna} search to determine the optimal XGBoost hyperparameters (Table \ref{tab:xgboost_classification_params}) tracking the best binary classification F1 scores. The best models for each property reached F1 scores of: 0.58, 0.71, and 0.68 on the validation sets accordingly.

\begin{table}[ht]
\centering
\caption{XGBoost Hyperparameters for Classification}
\label{tab:xgboost_classification_params}
\vskip 0.05in
\begin{tabular}{@{}ll@{}}
\toprule
\textbf{Hyperparameter}       & \textbf{Value/Range}          \\ \midrule
Objective                     & \texttt{binary:logistic}      \\
Lambda                        & \([1\text{e}{-8}, 10.0]\)     \\
Alpha                         & \([1\text{e}{-8}, 10.0]\)     \\
Colsample by Tree             & \([0.1, 1.0]\)               \\
Subsample                     & \([0.1, 1.0]\)               \\
Learning Rate                 & \([0.01, 0.3]\)              \\
Max Depth                     & \([2, 30]\)                  \\
Min Child Weight              & \([1, 20]\)                  \\
Tree Method                   & \texttt{hist}                \\
\bottomrule
\end{tabular}
\end{table}

\subsection{Binding Affinity Score Model}
We developed an unpooled reciprocal attention transformer model to predict protein-peptide binding affinity, leveraging latent representations from the ESM-2 650M protein language model \cite{lin2023evolutionary}. Instead of relying on pooled representations, the model retains unpooled token-level embeddings from ESM-2, which are passed through convolutional layers followed by cross-attention layers. The binding affinity data was split into a 0.8/0.2 ratio, maintaining similar affinity score distributions across splits. We used OPTUNA \cite{akiba2019optuna} for hyperparameter optimization tracing validation correlation scores. The final model was trained for 50 epochs with a learning rate of 3.84e-5, a dropout rate of 0.15, 3 initial CNN kernel layers (dimension 384), 4 cross-attention layers (dimension 2048), and a shared prediction head (dimension 1024) in the end. The classifier reached 0.64 Spearman's correlation score on validation data.

\subsection{Half-Life Score Model}
\textbf{Dataset Curation.} The half-life dataset is curated from three publicly available datasets: PEPLife, PepTherDia, and THPdb2 \citep{mathur2016peplife,d2021peptherdia,jain2024thpdb2}. Data related to human subjects were selected, and entries with missing half-life values were excluded. After removing duplicates, the final dataset consists of 105 entries.

\textbf{Pre-training on stability data.} Given the small size of the half-life dataset, which is insufficient for training a model to capture the underlying data distribution, we first pre-trained a score model on a larger stability dataset to predict peptide stability \citep{tsuboyama2023mega}. The model consists of three linear layers with ReLU activation functions, and a dropout rate of 0.3 was applied. The model was trained on a 2xH100 NVIDIA NVL GPU system with 94 GB of VRAM for 50 epochs. The Adam optimizer was employed with a learning rate 1e-2. A learning rate scheduler with 5 warm-up epochs and cosine decay was used, with initial and minimum learning rates both 1e-3. After training, the model achieved a validation Spearman's correlation of 0.7915 and an $R^2$ value of 0.6864, demonstrating the reliability of the stability score model.

\textbf{Fine-tuning on half-life data.} The pre-trained stability score model was subsequently fine-tuned on the half-life dataset. Since half-life values span a wide range, the model was adapted to predict the base-10 logarithm of the half-life (h) values to stabilize the learning process. After fine-tuning, the model achieved a validation Spearman's correlation of 0.8581 and an $R^2$ value of 0.5977.

\section{Sampling Details}
\subsection{Peptide Binder Generation Tasks}
\textbf{Score Model Settings.} To align all objectives as maximization, we convert the predicted hemolysis rate $h$ into a score $1-h$, so that lower hemolysis yields a higher value. We also cap the predicted log-scale half-life at 2 (i.e., 100 hours) to prevent it from dominating the optimization and ensure balanced trade-offs across all properties. For the remaining objectives—non-fouling, solubility, and binding affinity—we directly employ their model outputs during sampling.

\textbf{Hyperparameter Settings.} The hyperparameters were set as follows: The number of divisions used in generating weight vectors, $\text{num\_div}$, was set to 64, $\lambda$ to 1.0,  $\beta$ to 1.0, $\alpha_r$ to 0.5, $\tau$ to 0.3, $\eta$ to 1.0, $\Phi_{init}$ to $45^{\circ}$, $\Phi_{min}$ to $15^{\circ}$, $\Phi_{max}$ to $75^{\circ}$. The total sampling step $T$ was 100. 

\textbf{Importance Vectors.} In the task with five property guidance, the importance vector was set to $[1,1,1,0.5,0.2]$, each corresponding to hemolysis, non-fouling, solubility, half-life, and binding affinity guidance, respectively. For the two tasks with only three property guidance, the importance vector was set to $[1, 1, 0.1]$ for solubility, hemolysis, and binding affinity guidance, respectively, and $[1, 0.5, 0.2]$ for non-fouling, half-life, and binding affinity guidance, respectively. The rationale for setting the importance values is based on the range lengths of the properties: hemolysis, non-fouling, and solubility each have a range length of 1.0, half-life has a range length of 2.0, and binding affinity has a range length of 10.0. The importance values were assigned inversely proportional to these range lengths.

\subsection{Enhancer DNA Generation Tasks}
\textbf{Hyperparameter Settings.} The hyperparameters were set the same as those in peptide binder generation tasks, except that the total sampling step $T$ was set to 800.

\textbf{Importance Vectors.} The importance vector was set to be [1, 10] for the first task and [1, 100] for the second task, with the first value corresponding to the enhancer class guidance and the second value corresponding to the DNA shape guidance. The rationale for assigning these importance values is based on the range lengths of the properties: enhancer class probability has a range length of 1.0, HelT shape values have a range length of 2.0, and Rise shape values have a range length of 0.1. The importance values were assigned inversely proportional to these range lengths.

\section{Additional Proof} \label{proof}

\textbf{Claim:} \textsc{MOG-DFM} directs the discrete generation process toward the Pareto front by inducing a positive expected improvement in the direction of a specified weight vector \( \omega \in \mathbb{R}^N \).

\textbf{Proof:}  
Let \( \mathcal{S} = \mathcal{T}^d \) be the discrete sequence space over vocabulary \( \mathcal{T} \), and let \( x \in \mathcal{S} \) denote the current sequence state at time \( t \in [0,1] \). Assume the multi-objective score function \( s: \mathcal{S} \rightarrow \mathbb{R}^N \) is measurable, with \( N \) scalar objectives. Define the improvement vector at a candidate transition \( y^i \in \mathcal{T} \setminus \{x^i\} \) at position \( i \in \{1, \ldots, d\} \) as:
\[
\Delta s(y^i, x) := s(x^{(i \rightarrow y^i)}) - s(x),
\]
where \( x^{(i \rightarrow y^i)} \) denotes the sequence \( x \) with token \( x^i \) replaced by \( y^i \).

Let \( \omega \in \mathbb{R}^N \) be a fixed unit-norm trade-off vector sampled uniformly from the Das–Dennis lattice covering the simplex \( \Delta^{N-1} \). Define the directional improvement of a transition \( y^i \) as:
\[
D(y^i, x; \omega) := \Delta s(y^i, x) \cdot \omega.
\]

Define the set of feasible transitions (those within the hypercone of angle \( \Phi \in (0, \pi) \)) at time \( t \) as:
\[
Y^i(x, \omega, \Phi) := \left\{ y^i \in \mathcal{T} \setminus \{x^i\} \;\middle|\; \arccos\left( \frac{\Delta s(y^i, x) \cdot \omega}{\|\Delta s(y^i, x)\| \cdot \|\omega\|} \right) \leq \Phi \right\}.
\]

Let \( \mu^i_t(\cdot \mid x, \omega) \) be the conditional probability measure over feasible transitions defined by:
\[
\mu^i_t(y^i \mid x, \omega) := \frac{\exp\left( \Delta S(y^i, x, \omega) \right)}{Z(x, \omega)} \cdot \mathbf{1}_{\{ y^i \in Y^i(x, \omega, \Phi) \}},
\]
where \( \Delta S(\cdot) \) is the rank-directional guidance score and \( Z(x, \omega) := \sum_{y^i \in Y^i} \exp\left( \Delta S(y^i, x, \omega) \right) \) is the normalizing partition function. Assume that \( Y^i(x, \omega, \Phi) \) is non-empty, or else the algorithm falls back to selecting the best \( y^i \) with \( D(y^i, x; \omega) > 0 \) by construction.

We now consider the expected improvement in the direction of \( \omega \) over all guided transitions:
\[
\mathbb{E}_{i \sim \mathcal{U}[d], \; y^i \sim \mu^i_t(\cdot \mid x, \omega)} \left[ D(y^i, x; \omega) \right]
= \frac{1}{d} \sum_{i=1}^d \sum_{y^i \in Y^i(x, \omega, \Phi)} D(y^i, x; \omega) \cdot \mu^i_t(y^i \mid x, \omega).
\]

Since each \( y^i \in Y^i(x, \omega, \Phi) \) satisfies \( \arccos\left( \frac{\Delta s(y^i, x) \cdot \omega}{\|\Delta s(y^i, x)\| \cdot \|\omega\|} \right) \leq \Phi < \pi \), it follows that \( D(y^i, x; \omega) > 0 \) for all \( y^i \in Y^i \). Moreover, \( \mu^i_t(y^i \mid x, \omega) > 0 \) by construction.

Therefore, each term in the sum is strictly positive, and thus:
\[
\mathbb{E}[\Delta s(x_{\text{new}}, x) \cdot \omega] > 0,
\]
where \( x_{\text{new}} = x^{(i \rightarrow y^i)} \) is the updated sequence following a guided and filtered transition.

Hence, the MOG-DFM procedure ensures that in expectation, the sampling dynamics induce forward motion along the Pareto trade-off direction \( \omega \), thereby steering generation toward the Pareto frontier.

\hfill \qedsymbol

\newpage
\begin{table}[ht]
\centering
\caption{Average solubility (LogS) and half-life (in hours) metrics computed by ADMET-AI for each target across the 100 MOG-DFM-designed binders.}
\vspace{0.5em}
\small
\begin{tabular}{c|cc}
\hline
\textbf{Target}  & \textbf{LogS} & \textbf{Half-Life}\\
\hline
AMHR2       & -2.3931 & 15.505 \\
AMHR2       & -2.5055 & 18.777 \\
AMHR2       & -2.5784 & 16.463 \\
EWS::FLI1    & -2.3869 & 18.945 \\
EWS::FLI1    & -2.3813 & 16.305 \\
EWS::FLI1    & -2.5457 & 15.984 \\
MYC         & -2.4053 & 16.491 \\
OX1R        & -2.4772 & 23.002 \\
DUSP12      & -2.4333 & 19.258 \\
 1B8Q
& -2.3203&18.7862\\
 1E6I
& -2.0394&19.9358
\\
3IDJ
& -2.4193& 20.3586
\\
5AZ8
& -2.5964& 16.3016
\\
7JVS& -2.4824& 20.2565\\
\hline
\end{tabular}

\label{tab:admet_ai}
\end{table}

\begin{table}[ht]
\midfootnotesize
\centering
\caption{\textbf{Ablation study results for the adaptive hypercone filtering module in MOG-DFM.} Three settings are evaluated: 'w/o filtering' indicates the module is completely disabled, 'w/o adaptation' means the module is enabled but the hypercone is not adaptive, and 'MOG-DFM' represents the complete algorithm. For each setting, 100 peptide binders were designed, with lengths of 7, 12, and 12 for the targets 3IDJ, 4E-BP2, and EWS::FLI1, respectively.}
\vspace{0.5em}
\label{tab:algorithm_ablation}
\begin{tabular}{cc|ccccc}
\hline
Target & Method & Hemolysis ($\downarrow$) & Non‑Fouling& Solubility& Half‑Life& Affinity\\
\hline
\multirow{3}{*}{3IDJ}
  & w/o filtering   & 0.0660 & 0.8430 & 0.8482 & 12.50 & 7.3730 \\
  & w/o adaptation  & 0.0856 & 0.8060 & 0.7970 & 37.17 & 7.3142 \\
  & MOG‑DFM         & 0.0924 & 0.8246 & 0.7992 & 30.39 & 7.6304 \\
\hline
\multirow{3}{*}{4E-BP2}
  & w/o filtering   & 0.0504 & 0.8582 & 0.8600 & 12.62 & 6.5066 \\
  & w/o adaptation  & 0.0638 & 0.8418 & 0.8234 & 23.44 & 6.4548 \\
  & MOG‑DFM         & 0.0698 & 0.8210 & 0.8050 & 34.88 & 6.5824 \\
\hline
\multirow{3}{*}{EWS::FLI1}
  & w/o filtering   & 0.0450 & 0.8596 & 0.8570 &  4.40 & 6.1392 \\
  & w/o adaptation  & 0.0620 & 0.8444 & 0.8482 & 28.82 & 6.2118 \\
  & MOG‑DFM         & 0.0616& 0.8302& 0.8130& 34.225& 6.3631\\
\hline
\end{tabular}
\end{table}

\renewcommand{\arraystretch}{1.5}
\begin{table}[ht]
\centering
\caption{Ablation results for peptide binder design targeting PDB 7LUL with different guidance settings. For each setting, 100 binders of length 7 were designed.}
\vspace{0.5em}
\begin{tabular}{l|ccc}
\hline
\textbf{\makecell[c]{Guidance Settings\\[0.4ex]Affinity\hspace{0.5em}Solubility\hspace{0.5em}Hemolysis}} & \textbf{Affinity} & \textbf{Solubility} & \textbf{Hemolysis ($\downarrow$)} \rule{0pt}{1.6em} \\ 
\hline
\makecell[l]{\hspace{1em}$\checkmark$ \hspace{3.2em} $\checkmark$ \hspace{3.2em} $\times$} & 6.3489 & 0.8890 & 0.0620 \\
\hline
\makecell[l]{\hspace{1em}$\times$ \hspace{3.2em} $\checkmark$ \hspace{3.2em} $\checkmark$} & 5.0514 & 0.9482 & 0.0406 \\
\hline
\makecell[l]{\hspace{1em}$\checkmark$ \hspace{3.2em} $\times$ \hspace{3.2em} $\checkmark$} & 6.9060 & 0.4224 & 0.0488 \\
\hline
\makecell[l]{\hspace{1em}$\checkmark$ \hspace{3.2em} $\checkmark$ \hspace{3.2em} $\times$} & 6.5304 & 0.8975 & 0.1019 \\
\hline
\makecell[l]{\hspace{1em}$\times$ \hspace{3.2em} $\times$ \hspace{3.2em} $\checkmark$} & 5.0761 & 0.7148 & 0.0163 \\
\hline
\makecell[l]{\hspace{1em}$\times$ \hspace{3.2em} $\checkmark$ \hspace{3.2em} $\times$} & 5.2434 & 0.9772 & 0.0955 \\
\hline
\makecell[l]{\hspace{1em}$\checkmark$ \hspace{3.2em} $\times$ \hspace{3.2em} $\times$} & 7.4834 & 0.1218 & 0.3281 \\
\hline
\makecell[l]{\hspace{1em}$\times$ \hspace{3.2em} $\times$ \hspace{3.2em} $\times$} & 5.5631 & 0.3736 & 0.1567 \\
\hline
\end{tabular}

\label{tab:ablation_7LUL}
\end{table}
\renewcommand{\arraystretch}{1.5}
\begin{table}[ht]
\centering
\caption{Ablation results for peptide binder design targeting PDB CLK1 with different guidance settings. For each setting, 100 binders of length 12 were designed.}
\vspace{0.5em}
\begin{tabular}{l|ccc}
\hline
\textbf{\makecell[c]{Guidance Settings\\[0.4ex]Affinity\hspace{0.5em}Non-Fouling\hspace{0.5em}Half-Life}} 
  & \textbf{Affinity} 
  & \textbf{Non‑Fouling} 
  & \textbf{Half‑Life} \rule{0pt}{1.6em} \\
\hline
\makecell[l]{\hspace{1em} $\checkmark$ \hspace{3.8em}
             $\checkmark$ \hspace{3.8em}
            $\checkmark$}& 6.9194 & 0.7401 & 51.73 \\
\hline
\makecell[l]{\hspace{1em} $\times$ \hspace{3.8em}
            $\checkmark$ \hspace{3.8em}
            $\checkmark$}& 6.4735 & 0.8107 & 60.75 \\
\hline
\makecell[l]{\hspace{1em} $\checkmark$ \hspace{3.8em}
             $\times$ \hspace{3.8em}
             $\checkmark$}& 7.5360 & 0.3062 & 84.70 \\
\hline
\makecell[l]{\hspace{1em} $\checkmark$ \hspace{3.8em}
             $\checkmark$ \hspace{3.8em} 
             $\times$}
  & 7.4150 & 0.8560 &  1.24 \\
\hline
\makecell[l]{\hspace{1em} $\times$ \hspace{3.8em}
             $\times$ \hspace{3.8em} 
             $\checkmark$}
  & 6.2363 & 0.2624 & 96.44 \\
\hline
\makecell[l]{\hspace{1em} $\times$ \hspace{3.8em}
             $\checkmark$ \hspace{3.8em} 
             $\times$}
  & 6.1378 & 0.9503 &  0.94 \\
\hline
\makecell[l]{\hspace{1em} $\checkmark$ \hspace{3.8em}
             $\times$ \hspace{3.8em} 
             $\times$}
  & 8.5943 & 0.2439 &  3.15 \\
\hline
\makecell[l]{\hspace{1em} $\times$ \hspace{3.8em}
             $\times$ \hspace{3.8em} 
             $\times$}
  & 5.8926 & 0.3999 &  1.94 \\ 
\hline

\end{tabular}
\label{tab:ablation_CLK1}
\end{table}

\renewcommand{\arraystretch}{1.3}
\begin{table}[ht]
\centering
\midfootnotesize
\caption{Hyperparameter sensitivity benchmark for MOG-DFM in peptide binder generation, guided by five objectives. For each setting, 100 peptide binders are designed with a length matching that of the pre-existing binder for each target.}
\vspace{0.5em}
\label{tab:hyperparam_sensitivity}
\begin{tabular}{c|c|c|ccccc}
\hline
\textbf{\makecell{Hyper\\parameter}} 
  & \textbf{Target}
  & \textbf{Value} 
  & \textbf{Hemolysis} ($\downarrow$) 
  & \textbf{Non‑Fouling}& \textbf{Solubility}& \textbf{Half‑Life}& \textbf{Affinity}\\
\hline
\multirow{3}{*}{$\text{num\_div}$}  
  & \multirow{3}{*}{6MLC}
  & 32  & 0.0994 & 0.8088 & 0.7924 & 38.39 & 6.5436 \\
  & 
  & 64  & 0.0863 & 0.8280 & 0.8232 & 34.91 & 6.3260 \\
  & 
  & 128 & 0.0890 & 0.8438 & 0.8386 & 32.97 & 6.4197 \\
\hline
\multirow{4}{*}{$\beta$}& \multirow{4}{*}{4IU7}& 0.5 & 0.0829& 0.7894& 0.761& 28.10& 6.7884\\
  & 
  & 1   & 0.0684& 0.8388& 0.8321& 41.78& 7.0002\\
 & & 1.5& 0.0585& 0.8588& 0.8582& 47.65&7.0505\\
  & 
  & 2   & 0.0615& 0.8461& 0.8416& 53.45& 7.0169\\
\hline
\multirow{3}{*}{$\lambda$}  
  & \multirow{3}{*}{1AYC}
  & 0.5 & 0.0703 & 0.8168 & 0.8152 & 30.89 & 6.4838 \\
  & 
  & 1   & 0.0647 & 0.8362 & 0.8207 & 33.28 & 6.4549 \\
  & 
  & 2   & 0.0587 & 0.8690 & 0.8461 & 41.90 & 6.5317 \\
\hline
\multirow{5}{*}{$\alpha_r$}  
  & \multirow{5}{*}{2Q8Y}
  & 0.1 & 0.0777 & 0.8361 & 0.8051 & 37.83 & 6.0569 \\
  & 
  & 0.3 & 0.0718 & 0.8441 & 0.8280 & 38.83 & 6.0484 \\
  & 
  & 0.5 & 0.0718 & 0.8529 & 0.8421 & 31.45 & 6.0445 \\
  & 
  & 0.7 & 0.0688 & 0.8403 & 0.8377 & 35.50 & 6.0839 \\
  & 
  & 0.9 & 0.0813 & 0.8288 & 0.8091 & 45.25 & 6.1599 \\
\hline
\multirow{3}{*}{$\eta$}  
  & \multirow{3}{*}{2LTV}
  & 0.5 & 0.0633 & 0.8437 & 0.8368 & 29.48 & 7.3657 \\
  & 
  & 1   & 0.0601 & 0.8256 & 0.8144 & 24.47 & 7.3111 \\
  & 
  & 2   & 0.0624 & 0.8125 & 0.7887 & 35.13 & 7.1974 \\
\hline
\multirow{5}{*}{$\Phi_{\!init}$}  
  & \multirow{5}{*}{5M02}
  & 15  & 0.0746 & 0.8285 & 0.8007 & 34.04 & 7.0335 \\
  & 
  & 30  & 0.0792 & 0.8393 & 0.8187 & 35.60 & 7.0251 \\
  & 
  & 45  & 0.0747 & 0.8338 & 0.8192 & 36.29 & 7.0944 \\
  & 
  & 60  & 0.0813 & 0.8095 & 0.7970 & 38.25 & 7.0932 \\
  & 
  & 75  & 0.0830 & 0.8139 & 0.7949 & 33.29 & 7.1261 \\
\hline
\multirow{3}{*}{$[\Phi_{\min},\Phi_{\max}]$}  
  & \multirow{3}{*}{3EQS}
  & [0,90]  & 0.0572 & 0.8385 & 0.8200 & 26.64 & 8.2201 \\
  & 
  & [15,75] & 0.0599 & 0.8373 & 0.8116 & 29.56 & 8.1673 \\
  & 
  & [30,60] & 0.0614 & 0.8159 & 0.8020 & 35.71 & 8.2313 \\
\hline
\multirow{5}{*}{$\tau$}  
  & \multirow{5}{*}{5E1C}
  & 0   & 0.0614 & 0.8252 & 0.8119 & 24.57 & 7.0112 \\
  & 
  & 0.1 & 0.0650 & 0.8017 & 0.7835 & 31.19 & 7.1067 \\
  & 
  & 0.3 & 0.0595 & 0.8224 & 0.8088 & 28.72 & 7.0756 \\
  & 
  & 0.5 & 0.0555 & 0.8310 & 0.8043 & 24.03 & 7.0862 \\
  & 
  & 0.7 & 0.0590 & 0.8360 & 0.8078 & 28.27 & 7.0477 \\
\hline
\multirow{4}{*}{$T$}& \multirow{4}{*}{5KRI}
  & 50  & 0.0757 & 0.7386 & 0.7219 & 15.22 & 6.9155 \\
  & 
  & 100 & 0.0580 & 0.8617 & 0.8504 & 30.25 & 6.9946 \\
  & 
  & 200 & 0.0525 & 0.8695 & 0.8621 & 41.53 & 7.2166 \\
  & 
  & 500 & 0.0518 & 0.8799 & 0.8760 & 57.65 & 7.2172 \\
\hline
\multirow{5}{*}{\makecell{importance\\weights}}& \multirow{5}{*}{4EZN}
  & [1,1,1,0.5,0.2]& 0.0877 & 0.5735 & 0.5485 & 28.17 & 6.4190 \\
  & 
  & [1,1,1,0.5,0.1]& 0.0836 & 0.6003 & 0.5738 & 21.99 & 6.3409 \\
  & 
  & [1,1,1,1,0.1]& 0.0892 & 0.5549 & 0.5272 & 33.58 & 6.3844 \\
  & 
  & [1,1,1,1,0.2]& 0.0958 & 0.5939 & 0.5647 & 34.80 & 6.4281 \\
  & 
  & [1,1,1,1,1]& 0.0960 & 0.5377 & 0.5007 & 29.65 & 6.8613 \\
\hline
\end{tabular}
\end{table}

\begin{table}[ht]
\centering
\small
\caption{Motif clusters and associated properties of enhancer DNA sequences. In this paper, each class refers to its corresponding cluster ID.}
\vspace{0.5em}
\begin{tabular}{c|c|c|c}
\hline
Cluster ID & \# of explainable ASCAVs & Motif Annotation & \# of Motifs in the cluster \\
\hline
cluster\_1                  & 3278                                                            & ATF                                                          & 71                                                              \\
cluster\_2                  & 1041                                                            & CTCF                                                         & 85                                                              \\
cluster\_3                  & 2480                                                            & EBOX                                                         & 91                                                              \\
cluster\_4                  & 4011                                                            & AP1                                                          & 191                                                             \\
cluster\_5                  & 1165                                                            & RUNX                                                         & 37                                                              \\
cluster\_6                  & 789                                                             & SP                                                           & 20                                                              \\
cluster\_7                  & 1285                                                            & ETS                                                          & 33                                                              \\
cluster\_8                  & 544                                                             & TEAD                                                         & 9                                                               \\
cluster\_9                  & 1024                                                            & TFAP                                                         & 53                                                              \\
cluster\_10                 & 334                                                             & Other                                                        & 4                                                               \\
cluster\_11                 & 935                                                             & SOX                                                          & 17                                                              \\
cluster\_12                 & 1010                                                            & CTCFL                                                        & 16                                                              \\
cluster\_13                 & 696                                                             & GATA                                                         & 7                                                               \\
cluster\_14                 & 141                                                             & Other                                                        & 2                                                               \\
cluster\_15                 & 601                                                             & TEAD                                                         & 6                                                               \\
cluster\_16                 & 805                                                             & Other                                                        & 7                                                               \\
cluster\_17                 & 270                                                             & Other                                                        & 4                                                               \\
cluster\_18                 & 475                                                             & Other                                                        & 5                                                               \\
cluster\_19                 & 473                                                             & ZNF                                                          & 6                                                               \\
cluster\_20                 & 395                                                             & Other                                                        & 4                                                               \\
cluster\_21                 & 393                                                             & Other                                                        & 4                                                               \\
cluster\_22                 & 768                                                             & NRF                                                          & 8                                                               \\
cluster\_23                 & 214                                                             & Other                                                        & 2                                                               \\
cluster\_24                 & 336                                                             & Other                                                        & 2                                                               \\
cluster\_25                 & 375                                                             & Other                                                        & 3                                                               \\
cluster\_26                 & 215                                                             & Other                                                        & 2                                                               \\
cluster\_27                 & 234                                                             & Other                                                        & 2                                                               \\
cluster\_28                 & 354                                                             & Other                                                        & 3                                                               \\
cluster\_29                 & 210                                                             & Other                                                        & 2                                                               \\
cluster\_30                 & 200                                                             & Other                                                        & 2                                                               \\
cluster\_31                 & 218                                                             & Other                                                        & 2                                                               \\
cluster\_32                 & 415                                                             & Other                                                        & 2                                                               \\
cluster\_33                 & 387                                                             & SOX                                                          & 2                                                               \\
cluster\_34                 & 116                                                             & Other                                                        & 1                                                               \\
cluster\_35                 & 121                                                             & Other                                                        & 1                                                               \\
cluster\_36                 & 394                                                             & Other                                                        & 2                                                               \\
cluster\_37                 & 112                                                             & Other                                                        & 1                                                               \\
cluster\_38                 & 111                                                             & Other                                                        & 1                                                               \\
cluster\_39                 & 107                                                             & Other                                                        & 1                                                               \\
cluster\_40                 & 118                                                             & Other                                                        & 1                                                               \\
cluster\_41                 & 144                                                             & Other                                                        & 1                                                               \\
cluster\_42                 & 105                                                             & Other                                                        & 1                                                               \\
cluster\_43                 & 102                                                             & Other                                                        & 1                                                               \\
cluster\_44                 & 108                                                             & Other                                                        & 1                                                               \\
cluster\_45                 & 114                                                             & Other                                                        & 1                                                               \\
cluster\_46                 & 118                                                             & Other                                                        & 1                                                               \\
cluster\_47                 & 119                                                             & Other                                                        & 1                                                               \\
\hline
\end{tabular}
\label{tab:enhancer_data}
\end{table}
\newpage
\begin{figure}
    \centering
    \includegraphics[width=\textwidth]{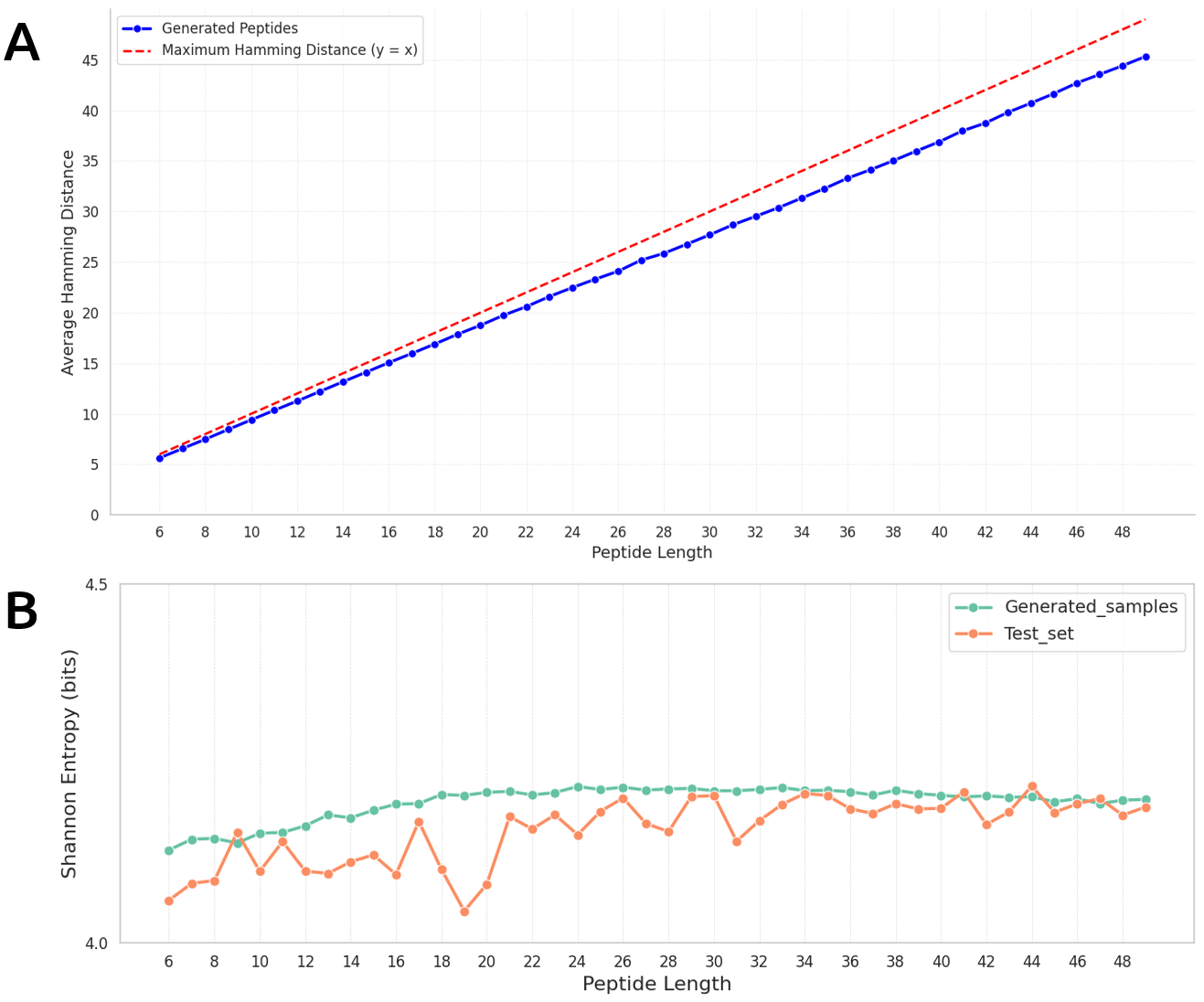}
    \caption{\textbf{(A)} The Hamming distance of sampled peptides of different lengths to the peptides of the same length in the test set.  \textbf{(B)} The Shannon Entropy of sampled peptides of different lengths to the peptides of the same length in the test set. }
    \label{fig:shannon}
\end{figure}
\begin{figure}
    \centering
    \includegraphics[width=\textwidth]{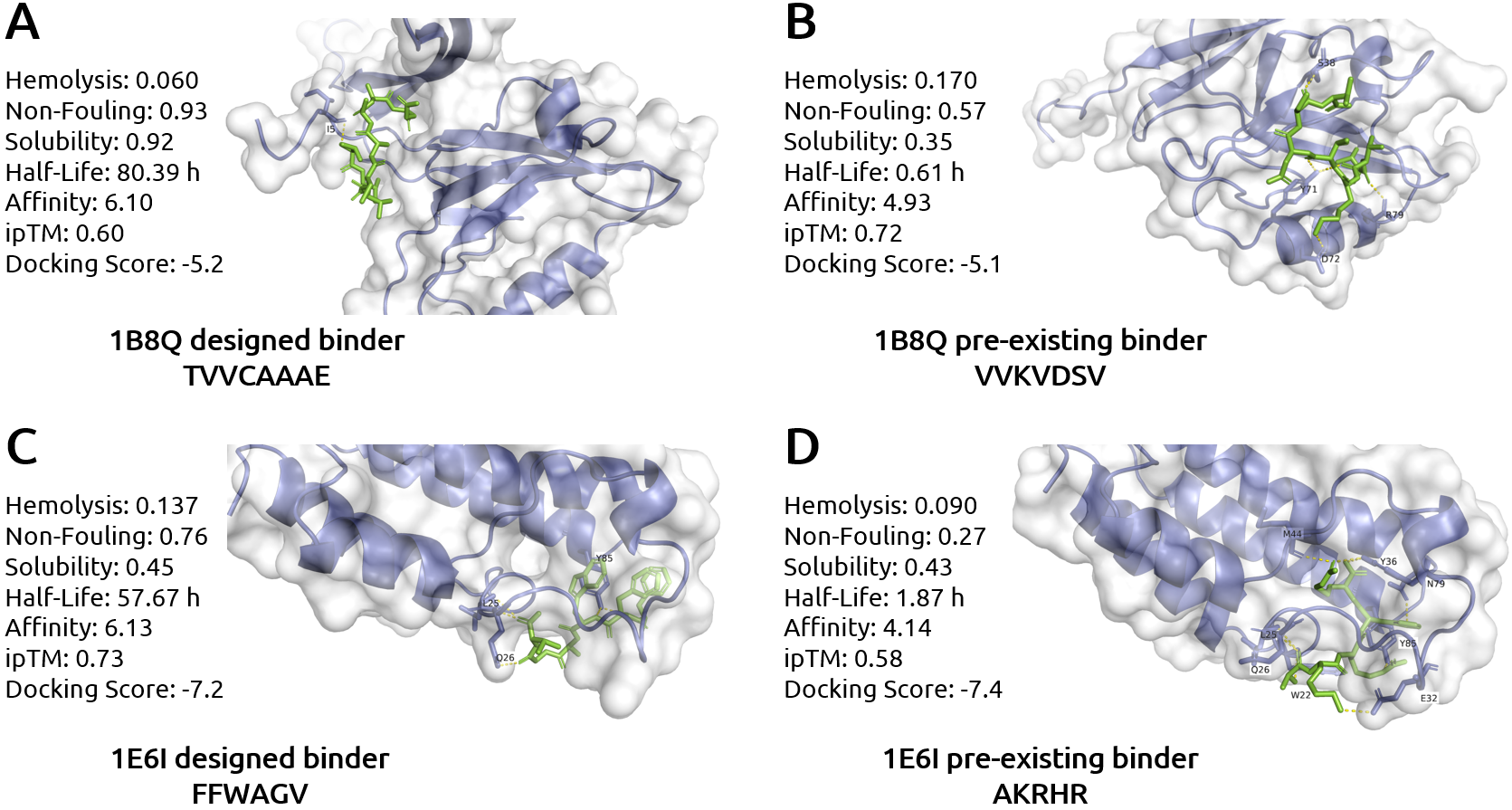}
    \par\vspace{0.7em}\par
    \includegraphics[width=\textwidth]{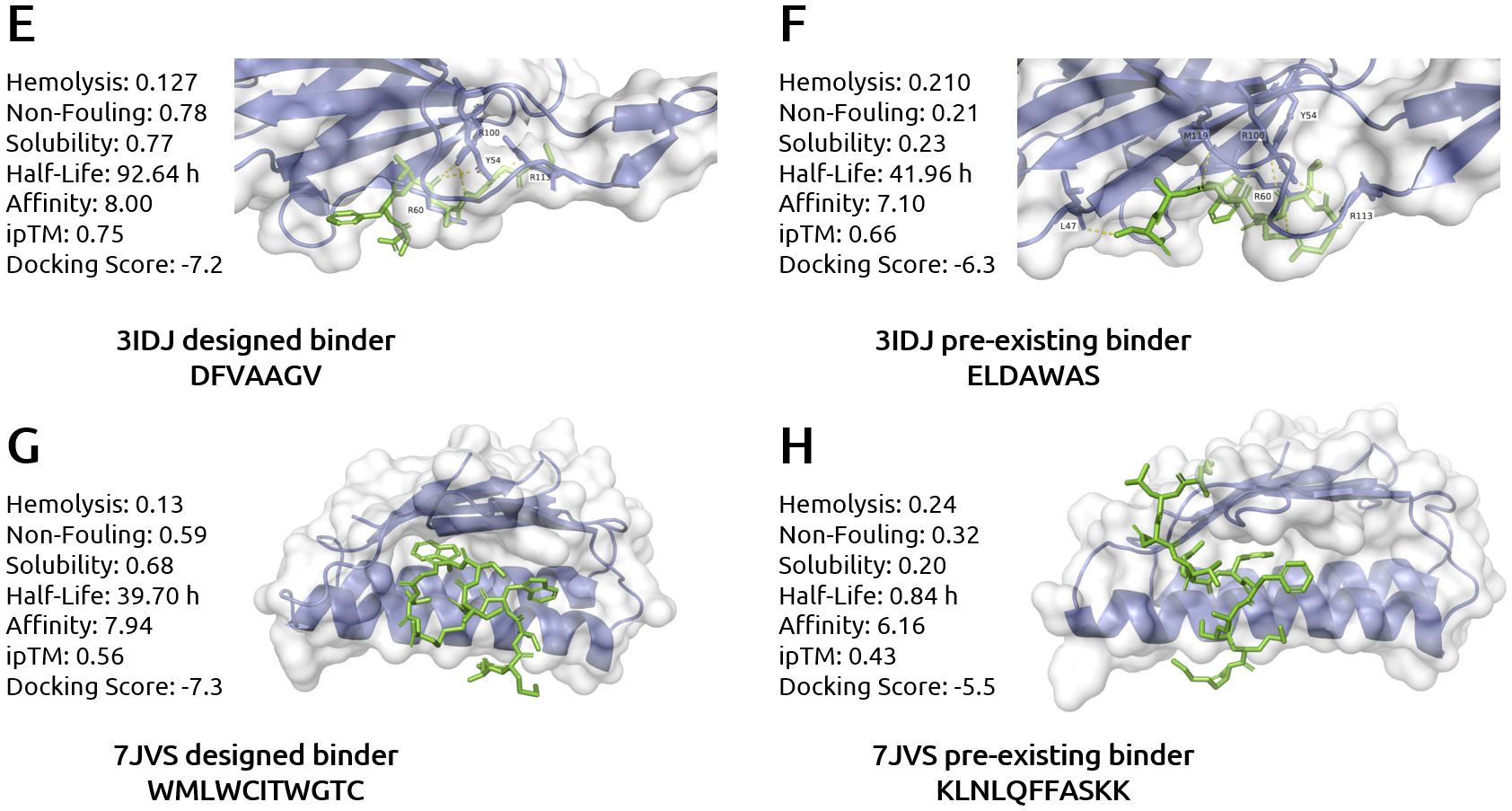}
    \caption{\textbf{Complex structures of target proteins with pre-existing binders.} \textbf{(A)-(B)} 1B8Q, \textbf{(C)-(D)} 1E6I, \textbf{(E)-(F)} 3IDJ, \textbf{(G)-(H)} 7JVS. Each panel shows the complex structure of the target with either a MOG-DFM-designed binder or its pre-existing binder. For each binder, five property scores are provided, as well as the ipTM score from AlphaFold3 and the docking score from AutoDock VINA. Interacting residues on the target are visualized.}
    \label{fig:w_binders}
\end{figure}
\begin{figure}
    \centering
    \includegraphics[width=\textwidth]{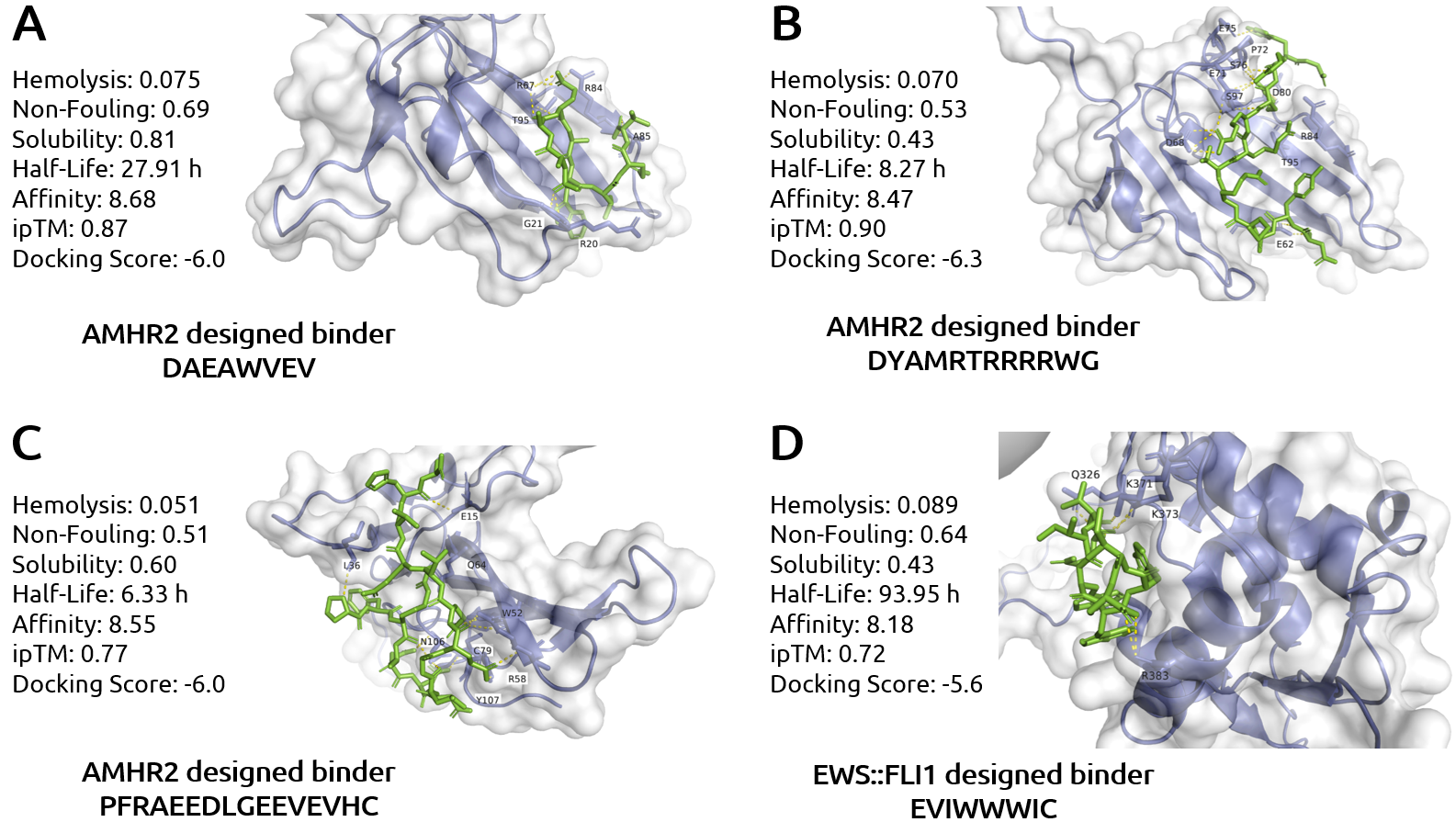}
    \par\vspace{0.7em}\par
    \includegraphics[width=\textwidth]{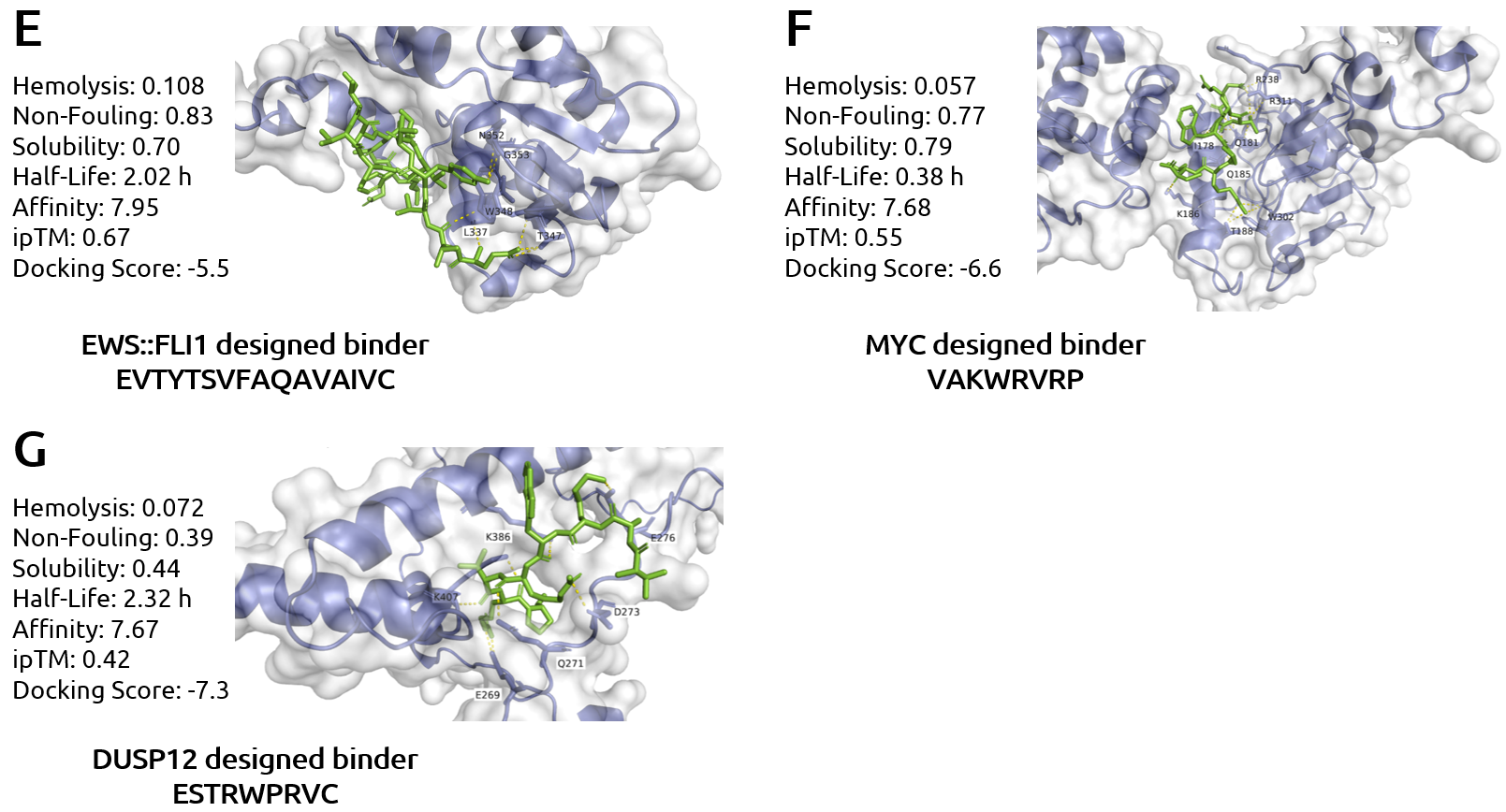}
    \caption{\textbf{Complex structures of target proteins without pre-existing binders.} \textbf{(A)-(C)} AMHR2, \textbf{(D)-(E)} EWS::FLI1, \textbf{(F)} MYC,  \textbf{(G)} DUSP12. Each panel shows the complex structure of the target with a MOG-DFM-designed binder. For each binder, five property scores are provided, as well as the ipTM score from AlphaFold3 and the docking score from AutoDock VINA. Interacting residues on the target are visualized.}
    \label{fig:w/o_binders}
\end{figure}

\begin{algorithm}
\caption{MOG-DFM: Multi-Objective-Guided Discrete Flow Matching}
\begin{algorithmic}[1]
\State \textbf{Input:} Pre-trained discrete flow matching model, multi-objective score functions
\State \textbf{Output:} Sequence $x_1$ with multi-objective optimized properties

\State \textbf{Initialize:} 
\State \quad Sample an initial sequence $x_0$ uniformly from the discrete state space $S$
\State \quad Generate a set of weight vectors $\{\omega_k\}_{k=1}^{M}$ that uniformly cover the N-dimensional Pareto front
\State \quad Select a weight vector $\omega$ randomly from $\{\omega_k\}$

\For{$t = 0$ to $1$ with step size $h = \frac{1}{T}$}
    \State \textbf{Step 1: Guided Transition Scoring}
    \State \quad Select a position $i$ in the sequence to update
    \State \quad For each candidate transition $y_i \neq x_i$:
    \State \quad \quad Compute the normalized rank score $I_n(y_i, x)$ for each objective $n$
    \State \quad \quad Compute $D(y_i, x, \omega)$ based on the alignment of improvements with $\omega$
    \State \quad \quad Combine rank and direction components: 
    \[
    \Delta S(y_i, x, \omega) = \text{Norm}\left[\frac{1}{N} \sum_{n=1}^{N} I_n(y_i, x)\right] + \lambda \cdot \text{Norm}\left[D(y_i, x, \omega)\right]
    \]
    \State \quad Re-weight the original velocity field $u_i(y_i, x)$ by the combined score

    \State \textbf{Step 2: Adaptive Hypercone Filtering}
    \State \quad Compute angle $\alpha_i$ between improvement vector $\Delta s(y_i, x)$ and weight vector $\omega$
    \State \quad Accept transitions $y_i$ where $\alpha_i \leq \Phi$ (hypercone angle)
    \State \quad Select the best transition $y_i^{best}$ from the candidates
    \State \quad \textbf{Adapt Hypercone Angle:}
    \State \quad \quad Compute the rejection rate $r_t$ based on the number of rejected candidate transitions
    \State \quad \quad Compute the exponential moving average $\overline{r_t}$ of rejection rate
    \State \quad \quad Update the hypercone angle $\Phi$ based on the moving average:
    \[
    \Phi_{t+h} = \text{clip}\left( \Phi_t \exp\left( \eta \left( \overline{r_t} - \tau \right) \right), \Phi_{\min}, \Phi_{\max} \right)
    \]

    \State \textbf{Step 3: Euler Sampling}
    \State \quad Use Euler’s method to sample the next state based on the guided velocity field
    \State \quad Transition to the new sequence
    \State \quad Update time: $t \to t + h$
\EndFor
\State \textbf{Return:} Final sequence $x_1$
\end{algorithmic}
\label{algorithm}
\end{algorithm}
\clearpage
\newpage

\end{document}